\documentclass[lettersize,journal]{IEEEtran}
\usepackage{graphics} 
\usepackage{epsfig} 
\usepackage{mathptmx} 
\usepackage{times} 
\usepackage{amsmath} 
\usepackage{amssymb}  
\usepackage{multirow}
\usepackage{amsfonts}
\usepackage{booktabs}
\usepackage[table,xcdraw]{xcolor}
\usepackage{cite}
\usepackage{url}
\usepackage{threeparttable}
\usepackage{bigdelim}
\usepackage{indentfirst}
\usepackage{colortbl}
\usepackage{bbding}
\usepackage{subfiles}
\usepackage{soul}
\usepackage[colorlinks]{hyperref}
\hypersetup{
	colorlinks=true,
	linkcolor=red,
	filecolor=red,
	citecolor=blue,      
	urlcolor=cyan,
}
\soulregister{\ref}7
\soulregister{\cite}7

\definecolor{car}{rgb}{0.39215686, 0.58823529, 0.96078431}
\definecolor{bicycle}{rgb}{0.39215686, 0.90196078, 0.96078431}
\definecolor{motorcycle}{rgb}{0.11764706, 0.23529412, 0.58823529}
\definecolor{truck}{rgb}{0.31372549, 0.11764706, 0.70588235}
\definecolor{other-vehicle}{rgb}{0.39215686, 0.31372549, 0.98039216}
\definecolor{person}{rgb}{1., 0.11764706, 0.11764706}
\definecolor{bicyclist}{rgb}{1., 0.15686275, 0.78431373}
\definecolor{motorcyclist}{rgb}{0.58823529, 0.11764706, 0.35294118}
\definecolor{road}{rgb}{1., 0., 1.}
\definecolor{parking}{rgb}{1., 0.58823529, 1.}
\definecolor{sidewalk}{rgb}{0.29411765, 0., 0.29411765}
\definecolor{other-ground}{rgb}{0.68627451, 0., 0.29411765}
\definecolor{building}{rgb}{1., 0.78431373, 0.}
\definecolor{fence}{rgb}{1., 0.47058824, 0.19607843}
\definecolor{vegetation}{rgb}{0., 0.68627451, 0.}
\definecolor{trunk}{rgb}{0.52941176, 0.23529412, 0.}
\definecolor{terrain}{rgb}{0.58823529, 0.94117647, 0.31372549}
\definecolor{pole}{rgb}{1., 0.94117647, 0.58823529}
\definecolor{traffic-sign}{rgb}{1., 0., 0.}   

\definecolor{barrier}{RGB}{112,128,144}
\definecolor{bicycle}{RGB}{220,20,60}
\definecolor{bus}{RGB}{255, 127, 80}
\definecolor{car}{RGB}{255, 158, 0}
\definecolor{const. veh.}{RGB}{233, 150, 70}
\definecolor{motorcycle}{RGB}{255,61,99}
\definecolor{pedestrian}{RGB}{0,0,230}
\definecolor{traffic cone}{RGB}{47,79,79}
\definecolor{trailer}{RGB}{255,140,0}
\definecolor{truck}{RGB}{255,99,71}
\definecolor{drive. suf.}{RGB}{0,207,191}
\definecolor{other flat}{RGB}{175,0,75}
\definecolor{sidewalk}{RGB}{75,0,75}
\definecolor{terrain}{RGB}{112,180,60}
\definecolor{manmade}{RGB}{222,184,135}
\definecolor{vegetation}{RGB}{0,175,0}

\begin{document}

\title{
Camera-based 3D Semantic Scene Completion with \\ Sparse Guidance Network
}

\author{
Jianbiao Mei, Yu Yang, Mengmeng Wang, Junyu Zhu, Jongwon Ra, Yukai Ma, Laijian Li, and Yong Liu
\thanks{
Jianbiao Mei, Yu Yang, Mengmeng Wang, Junyu Zhu, Jongwon Ra, Yukai Ma, Laijian Li, and Yong Liu are with the Institute of Cyber-Systems and Control, Zhejiang University, Hangzhou, China (e-mail: jianbiaomei@zju.edu.cn; yu.yang@zju.edu.cn; mengmengwang@zju.edu.cn; junyuzhu@zju.edu.cn; jongwonra@zju.edu.cn; yukaima@zju.edu.cn; lilaijian@zju.edu.cn; yongliu@iipc.zju.edu.cn).}
\thanks{Corresponding Author: Yong Liu.}
\thanks{This work is supported by NSFC 62088101 Autonomous Intelligent Unmanned Systems.}}
\markboth{Journal of \LaTeX\ Class Files,~Vol.~14, No.~8, August~2023}%
{Shell \MakeLowercase{\textit{et al.}}: A Sample Article Using IEEEtran.cls for IEEE Journals}


\maketitle

\begin{abstract}
Semantic scene completion (SSC) aims to predict the semantic occupancy of each voxel in the entire 3D scene from limited observations, which is an emerging and critical task for autonomous driving. Recently, many studies have turned to camera-based SSC solutions due to the richer visual cues and cost-effectiveness of cameras. However, existing methods usually rely on sophisticated and heavy 3D models to process the lifted 3D features directly, which are not discriminative enough for clear segmentation boundaries. In this paper, we adopt the dense-sparse-dense design and propose a one-stage camera-based SSC framework, termed SGN, to propagate semantics from the semantic-aware seed voxels to the whole scene based on spatial geometry cues.  Firstly, to exploit depth-aware context and dynamically select sparse seed voxels, we redesign the sparse voxel proposal network to process points generated by depth prediction directly with the coarse-to-fine paradigm. Furthermore, by designing hybrid guidance (sparse semantic and geometry guidance) and effective voxel aggregation for spatial geometry cues, we enhance the feature separation between different categories and expedite the convergence of semantic propagation. Finally, we devise the multi-scale semantic propagation module for flexible receptive fields while reducing the computation resources. Extensive experimental results on the SemanticKITTI and SSCBench-KITTI-360 datasets demonstrate the superiority of our SGN over existing state-of-the-art methods. And even our lightweight version SGN-L achieves notable scores of 14.80\% mIoU and 45.45\% IoU on SeamnticKITTI validation with only 12.5 M parameters and 7.16 G training memory. Code is available at \url{https://github.com/Jieqianyu/SGN}.
\end{abstract}


\section{Introduction} \label{method:intro}
\IEEEPARstart{I}{n} recent years, there has been significant attention and rapid progress in 3D scene understanding, which constitutes the bedrock of autonomous driving systems and robotics. By precisely perceiving the occupancy and semantics of their surroundings, autonomous vehicles, and robotics can make informed decisions and navigate safely. To this end, Semantic Scene Completion (SSC)  has been introduced to predict the semantic occupancy of each voxel of the entire 3D scene from limited observation. SSC helps create a more comprehensive representation of the environment, which includes filling in the gaps or missing information in the sensor data. This can be essential for agents to identify obstacles, understand the road layout, and make safe decisions. However, accurately estimating the semantics and geometry of the real world from partial observations is challenging due to the complexities presented by real-world scenarios.

\begin{figure}[t]
\centering
	\includegraphics[width=0.49\textwidth]{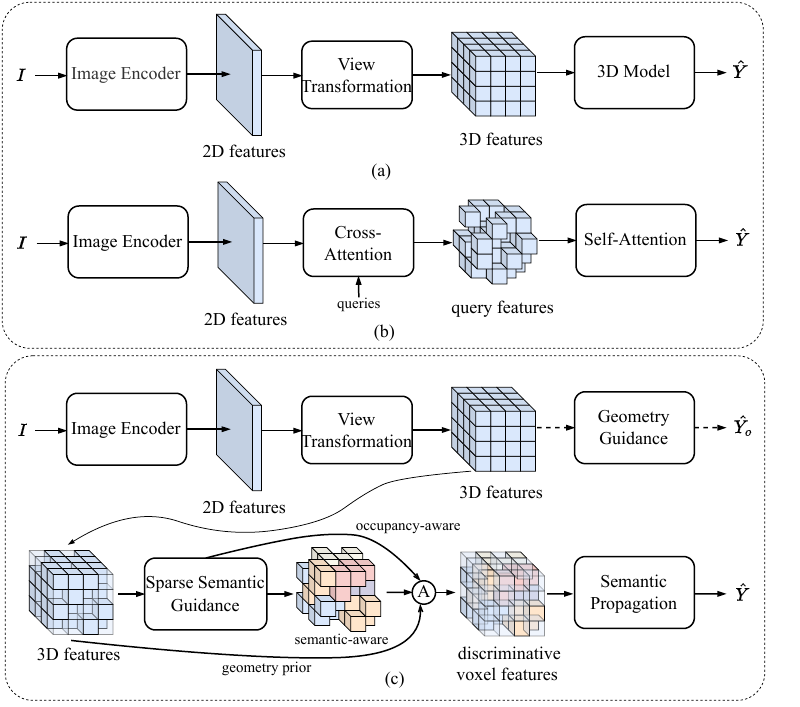}
	\caption{(a) Fully dense processing with heavy and complex 3D model. (b) MAE-like architecture in a ``sparse-to-dense" manner. (c) Our ``dense-sparse-dense" design with hybrid guidance and semantic propagation. ``A" denotes the voxel aggregation layer for geometry cues.}
	\label{fig:intro}
\end{figure}

SSC has attracted extensive studies due to its application prospects for downstream tasks such as mapping and planning. When working with outdoor driving scenes, LiDAR has emerged as a popular input modality for many existing methods \cite{roldao2020lmscnet, cheng2021s3cnet, rist2021semantic, yan2021sparse, xia2023scpnet, li2023lode} to capture 3D information of surroundings, but it suffers from high-cost sensors. Recently, there has been a shift towards camera-based SSC solutions. As the pioneer, MonoScene \cite{cao2022monoscene} proposed the first framework for monocular 3D SSC, utilizing mapping projection to lift RGB images to 3D volumes processed with the 3D UNet. Afterward, many camera-based methods such as OccDepth \cite{miao2023occdepth}, SurroundOcc \cite{wei2023surroundocc}, and OccFormer \cite{zhang2023occformer} are developed with a similar pipeline consisting of the image backbone, view transformer, and 3D model, as illustrated in Figure \ref{fig:intro} (a). However, they rely on sophisticated and heavy 3D models to process the lifted 3D features directly, which are not discriminative enough for clear segmentation boundaries. We explain that the lifted 3D features by 3D-2D mapping projection \cite{cao2022monoscene} contain many ambiguities due to the assumption of the uniform depth distribution and 2D-3D methods such as LSS \cite{philion2020lift} only utilize coarse surface information from depth distribution estimation. 

On the other hand, VoxFormer \cite{li2023voxformer} proposed an MAE-like architecture to complete non-visible structures using constructed visible areas. It adopts the two-stage framework, with the first stage for query proposal and the second stage for densification and segmentation. By completing the 3D scene in a \textbf{sparse-to-dense} manner shown in Figure \ref{fig:intro} (b), VoxFormer is more efficient and scalable than the dense processing with complicated 3D models mentioned above. However, it still suffers from several limitations. 
Firstly, the densification stage is mainly considered from the perspective of scene completion based on queries. The intra-category feature separation of queries is neglected. Besides, the second stage only considers the information from the queries that only include partial observation and are not always accurate, increasing the difficulty of subsequent completion and segmentation. Finally, the two-stage training and inference cannot fully consider global information due to the independent optimization of different stages. The geometry information from the first stage is also not fully utilized.  

To address the above problems, we propose a novel one-stage camera-based SSC framework, \textbf{S}parse \textbf{G}uidance \textbf{N}etwork (\textbf{SGN}), to propagate semantics from the semantic-aware seed voxels to the whole scene based on spatial geometry cues, as illustrated in Figure \ref{fig:intro} (c). Specifically, we employ the \textbf{dense-sparse-dense} design to implement the semantic propagation of semantic-aware seed features, avoiding relying on heavy 3D models to process coarse scene representations that are not discriminative enough.
Firstly, to dynamically select sparse seed voxels and encode depth-aware context, we redesign the sparse voxel proposal network to directly process points generated by depth prediction with the coarse-to-fine paradigm. 
And by further designing hybrid guidance (sparse semantic and geometry guidance) and effective voxel aggregation for spatial geometry cues, we enhance the intra-category feature separation and expedite the convergence of the semantic propagation. 
We also devise the multi-scale semantic propagation module using anisotropic convolutions \cite{li2020anisotropic} for flexible receptive fields while reducing the computation resources. By this means, our SGN is lightweight while having a more powerful representation ability.

Extensive experiments on the challenging SemanticKITTI \cite{behley2019semantickitti} and SSCBench-KITTI-360 \cite{li2023sscbench} datasets demonstrate the superiority of our SGN over existing state-of-the-art methods. For example, on the SemanticKITTI validation set, even our lightweight version SGN-L achieves notable scores of 14.80\% mIoU and 45.45\% IoU with only \textbf{12.5 M} parameters and \textbf{7.16 G} memory for training, exceeding VoxFormer-T by 1.45\% points in mIoU and 1.30\% points in IoU while being more lightweight and less memory consumption.

Our main contributions can be summarized as follows:
\begin{itemize}
    \item We propose a one-stage camera-based SSC framework called \textbf{SGN}, propagating semantics from the semantic- and occupancy-aware seed voxels to the whole scene based on spatial geometry cues.
    \item We adopt the \textbf{dense-sparse-dense} design and propose hybrid guidance and effective voxel aggregation to enhance intra-categories feature separation and expedite the convergence of the semantic propagation.
    \item Extensive experiments on the SemanticKITTI and SSCBench-KITTI-360 benchmarks demonstrate the effectiveness of our SGN, which is more lightweight and achieves the new state-of-the-art. 
\end{itemize}

\section{Related Works}
\paragraph{Semantic Scene Completion} Due to the vital application of semantic occupancy prediction in autonomous driving, SSC has attracted extensive attention. After the release of the large-scale outdoor benchmark SemanticKITTI \cite{behley2019semantickitti}, many outdoor SSC methods have emerged. According to the input modality, existing outdoor methods can be mainly classified into LiDAR-based and camera-based methods.

\textbf{LiDAR-based methods} consider LiDAR a primary modality to enable accurate 3D semantic occupancy prediction. Following the pioneering SSCNet \cite{song2017semantic}, UDNet \cite{zou2021up} exploits a single 3D U-Net framework to obtain predictions from the grids generated by the LiDAR points, resulting in extra computation overhead of empty voxels. Afterward, LMSCNet \cite{roldao2020lmscnet} introduces the 2D CNN for feature encoding, and SGCNet \cite{zhang2018efficient} uses spatial group convolutions to improve efficiency. 
Some solutions focus on multi-view fusion \cite{cheng2021s3cnet}, local implicit functions \cite{rist2021semantic}, and knowledge distillation \cite{xia2023scpnet} for SSC.
Besides, the relationships between semantic segmentation and scene completion are explored. For example, JS3C-Net \cite{yan2021sparse} and SSA-SC \cite{yang2021semantic} design a semantic segmentation network to assist the semantic scene completion. SSC-RS \cite{mei2023ssc} design multi-branch network to fuse semantic and geometry features hierarchically.

\begin{figure*}[t]
\centering
	\includegraphics[width=0.95\textwidth]{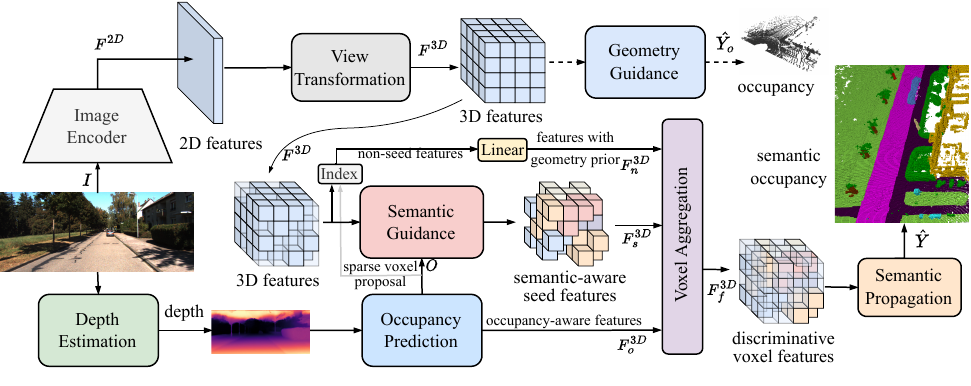}
	\caption{Overall framework of our SGN. The image encoder extracts 2D features, establishing the foundation for the 3D features generated through view transformation. An auxiliary occupancy head is applied to provide geometry guidance. The sparse semantic guidance consists of two parts: sparse voxel proposal and semantic guidance. The depth-based occupancy prediction is designed for the sparse voxel proposal. This proposal, along with the 3D features, is fed into the subsequent semantic guidance (depicted in Figure \ref{fig:sdb}) to index seed features and inject semantic context into these seed features. Afterward, the voxel aggregation layer combines the semantic-aware seed features, geometry prior from the non-seed features, and occupancy-aware features from the depth-based occupancy prediction. This forms the informative voxel features processed by the multi-scale semantic propagation for the final prediction.}
	\label{fig:pipeine}
\end{figure*}

\textbf{Camera-based methods.} Recently, camera-based perception such as detection \cite{huang2021bevdet, wang2022detr3d, liu2022petr, li2022bevformer, wang2021fcos3d, chen2022polar} and segmentation \cite{li2022bevformer, zhou2022cross, chen2022efficient} is currently more attractive due to cameras' richer visual cues and cost-effectiveness. 
And there is also a shift towards camera-based solutions \cite{cao2022monoscene, huang2023tri} to SSC. MonoScene \cite{cao2022monoscene} first proposed to infer 3D SSC from a single monocular RGB image, which applied a classical 3D UNet network to process the voxel features projected along the line of sight. Afterward, TPVFormer \cite{huang2023tri} proposed a tri-perspective view (TPV) representation to describe the fine-grained 3D structure of a scene. VoxFormer \cite{li2023voxformer} proposed an MAE-like architecture to complete non-visible structures using constructed visible areas. OccFormer \cite{zhang2023occformer} designed a dual-path transformer network. And SurroundOcc \cite{wei2023surroundocc} applied 3D convolutions to upsample multi-scale voxel features progressively and devised a pipeline to generate dense SSC ground truth. Symphonize \cite{jiang2024symphonize} modeled the scene volume with a sparse set of instance queries with context awareness. Some methods \cite{miao2023occdepth, li2023stereoscene} leveraged implicit stereo depth information and stereo matching to resolve geometry ambiguity. NDC-scene \cite{yao2023ndc} extends the 2D feature map to a Normalized Device Coordinates (NDC) space to alleviate the feature ambiguity, pose ambiguity and computation imbalance. Besides, there are some SSC solutions \cite{huang2023tri, wei2023surroundocc, gan2023simple, li2023fb} for multi-view cameras. And multiple benchmarks \cite{li2023sscbench, tian2023occ3d, wang2023openoccupancy} are proposed to facilitate the SSC's development.

We focus on camera-based SSC in outdoor scenarios. Compared with the existing works, our SGN proposes to propagate semantics from the semantic-aware seed voxels to the whole scene based on spatial geometry cues. SGN avoids relying on heavy and sophisticated 3D models to handle lifted voxel features with rough geometry context like many existing SSC methods \cite{cao2022monoscene, wei2023surroundocc, zhang2023occformer, miao2023occdepth, jiang2024symphonize}. Our method is built on the recent two-stage method VoxFormer \cite{li2023voxformer}. However, unlike VoxFormer, our SGN is one-stage, which adopts the dense-sparse-dense design and proposes hybrid guidance and effective voxel aggregation to enhance intra-categories feature separation and expedite the convergence of the semantic propagation.
Compared with VoxFormer, our SGN achieved better performance while being more lightweight and requiring less memory consumption.

\paragraph{Camera-based 3D Perception} Camera-based 3D perception, encompassing domains such as 3D detection \cite{huang2021bevdet, wang2022detr3d, liu2022petr, li2022bevformer, wang2021fcos3d, bao2019monofenet, huang2023obmo, chen2022polar} and segmentation \cite{li2022bevformer, zhou2022cross, chen2022efficient, cai2023ci3d, zhang20213d}, has gained increasing traction owing to the rich visual cues provided by cameras and their cost-effectiveness. Various monocular-based approaches have adapted 2D techniques to the 3D domain, such as FCOS3D \cite{wang2021fcos3d} and DETR3D \cite{wang2022detr3d}. In recent times, a significant shift has been observed in camera-based research toward Bird's Eye View (BEV) representations \cite{huang2021bevdet, li2022bevformer, li2023bevdepth, peng2023bevsegformer, zhang2022beverse, zhou2022cross, hu2021fiery, cai2023ci3d}, facilitated by view transformation techniques such as LSS \cite{philion2020lift}, OFT \cite{roddick2018orthographic}, and the cross-attention module \cite{li2022bevformer}. For example,  BEVDet \cite{huang2021bevdet} and BEVDepth \cite{li2023bevdepth} incorporate depth estimation to facilitate the transformation from perspective view to BEV. Additionally, BEVFormer \cite{li2022bevformer} employs cross-attention to inject cues from image features to BEV queries effectively.
The efficacy of BEV-based perception \cite{li2022bevformer, li2023bevdepth, zhang2022beverse, liu2022bevfusion, mei2023ssc, mei2023centerlps} has been validated by these advancements. However, for Semantic Scene Completion (SSC) tasks, the utilization of 3D voxel representations, which encapsulate more volumetric information, becomes imperative. As such, the quest to devise discriminative 3D scene representations and to process voxel features both effectively and efficiently remains a vibrant area of ongoing research and exploration.

\section{Method}
\subsection{Overview}
We show the overall framework of our SGN in Figure \ref{fig:pipeine}. SGN adopts the dense-sparse-dense design and propagates semantics from the semantic-aware seed voxels to the whole scene based on spatial geometry cues from the non-seed features and features from the depth-based occupancy prediction. SGN takes RGB images as the input and extracts 2D features using the image encoder. Then the 3D features are obtained through the view transformation. For dynamically indexing seed voxels, we generate the sparse voxel proposal based on depth prediction. Then according to the proposal and 3D features, the hybrid guidance is designed to inject semantic and geometry cues and facilitate feature learning. Furthermore, we develop the voxel aggregation layer to form the informative voxel features, which are processed by our multi-scale semantic propagation module for the final semantic occupancy prediction.

\paragraph{Image Encoder} We use ResNet-50 \cite{he2016deep} with FPN \cite{lin2017feature} to construct the image encoder for extracting 2D features from RGB images. The extracted features $\textbf{F}^{2D} \in \mathbb{R}^{N_t \times C\times H \times W}$ provide a strong foundation for the subsequent voxel features, where $N_t$ is the image number of temporal inputs, $C$ is the feature channel and $(H, W)$ denotes the image resolution.

\paragraph{View Transformation} Similar to MonoScene \cite{cao2022monoscene}, we construct 3D features by sampling 2D features via 3D-2D projection mapping with camera parameters.
The simple projection mapping operation provides coarse volumetric scene representation for the latter contextual modeling. And it is more convenient and concise than learnable LSS \cite{philion2020lift} and cross-attention mechanism \cite{li2022bevformer}. 

Let $\textbf{x} \in \mathbb{R}^{X \times Y \times Z \times 3}$ denote the centroid of $X \times Y \times Z$ voxels in world coordinates. We establish the projection mapping $\pi(\textbf{x})$ using the camera parameters $(\textbf{K}, \textbf{T})$, where $\textbf{K}$ and $\textbf{T}=[\textbf{R}, \textbf{t}]$ are the cameras intrinsic and extrinsic matrices directly provided in KITTI \cite{geiger2012we}. Let $p$ denotes a point in $\textbf{x}$, the mapping function establishes the relationship between the point and the image pixel $(u, v)$, which can be represented by: 
\begin{equation}
    [x_c, y_c, z_c]^T = \textbf{R}\cdot p + \textbf{t}
\end{equation} 
\begin{equation}
    z_c \circ [u, v, 1]^T = \textbf{K}\cdot[x_c, y_c, z_c]^T
\end{equation} where $\circ$ denote element-wise product. 

We take the average of sampled features from different images for each voxel. And the features of voxels outside the field of view (FOV) are set to zero.
Mathematically, the 3D features $\textbf{F}^{3D} \in \mathbb{R}^{C\times X \times Y \times Z}$ are sampled from the 2D features $\textbf{F}^{2D}$ as follows:
\begin{equation}
    \textbf{F}^{3D} = W \cdot \sum_{t=1}^{N_t}{[\phi_{\pi(\textbf{x})}(\textbf{F}_t^{2D})\cdot \textbf{M}_t^{FOV}]}
\end{equation}
\begin{equation}
    W_p=\left\{
    \begin{aligned}
    1/\delta_p & , & \delta_p > 0, \\
    1 & , & \delta_p = 0.
    \end{aligned}
    \right.
\end{equation}
where $\phi_a(b)$ is the sampling function that samples features $b$ at coordinates $a$, $\textbf{F}^{2D}_t$ is the 2D features of image $\textbf{I}_t$, $\textbf{M}^{FOV}_t \in \mathbb{R}^{1\times X \times Y \times Z}$ is the binary mask indicating the field of view of image $\textbf{I}_t$, $\delta_p$ is the number of hit images for point $p$ in $\textbf{x}$, $W_p$ is the weight value for points $p$ in $W$.

\subsection{Feature Learning with Hybrid Guidance}
As discussed above, most existing methods design heavy and complicated models to directly process the 3D features $\textbf{F}^{3D}$ produced by the view transformation module for the final semantic scene prediction. We argue that the coarse scene representation $\textbf{F}^{3D}$ is not discriminative enough to get clear segmentation boundaries, which slows down the convergence of the model. Therefore, we propose sparse semantic guidance and geometry guidance to inject semantic and geometry cues for informative voxel features.

\subsubsection{Geometry Guidance.}
We first attach the auxiliary 3D occupancy head as the geometry guidance after the 3D features from the view transformation module to provide coarse geometry awareness. Specifically, we construct the 3D occupancy head with an anisotropic convolution layer \cite{li2020anisotropic} and a linear layer. In the spirit of \cite{zhou2020cylinder3d}, the anisotropic convolution decomposes a 3D convolution operation into three consecutive 1D convolutions in different directions. Additionally, each of these 1D convolutions is equipped with a mixer containing distinct kernel sizes, enhancing the model's ability to learn and extract meaningful features from the input data. It can provide flexible receptive fields while alleviating resource consumption. By predicting the 3D occupancy $\hat{\textbf{Y}}_o$ on the lifted 3D features $\textbf{F}^{3D}$ using the auxiliary head, we apply the guidance on the coarse scene representation and provide the geometry prior for the latter seed features' semantic prediction and propagation. We optimize the occupancy probability with binary cross-entropy loss:
\begin{equation}
    \mathcal{L}_{geo} = -\sum_{i}[(1-\textbf{Y}_{o, i})\mathrm{log}(1-\hat{\textbf{Y}}_{o, i}) + \textbf{Y}_{o, i}\mathrm{log}(\hat{\textbf{Y}}_{o, i})]
\end{equation} where $i$ indexs the voxel of the 3D scene and $\textbf{Y}_{o}$ is the occupancy ground truth. Note that the auxiliary 3D head is abandoned during inference and using geometry guidance does not introduce any extra computation.

\begin{figure}[t]
\centering
	\includegraphics[width=0.49\textwidth]{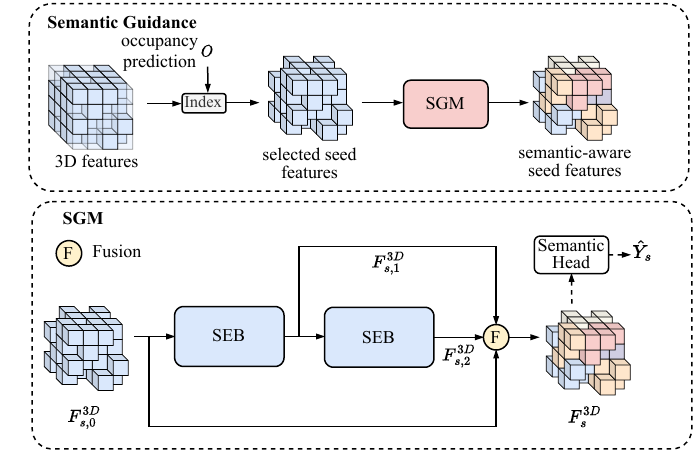}
	\caption{Detailed architecture of the proposed semantic guidance module (SGM). The sparse encoder block (SEB) consists of a sparse feature encoder and a sparse geometry feature encoder adopted from \cite{ye2022efficient}.}
	\label{fig:sdb}
\end{figure}

\subsubsection{Sparse Semantic Guidance.} Since directly learning the semantics of all the voxels from the 3D features with coarse volumetric information is less effective and efficient, we propose propagating \textbf{semantics} from \textbf{seed} voxel to the whole scene. Specifically, we generate the sparse voxel proposal to choose seed voxels and encourage inter-category separability of seed features with semantic guidance, expediting the latter semantic propagation.

\paragraph{Sparse Voxel Proposal} We devise the sparse voxel proposal network (SVPN) to generate the sparse proposal for indexing seed voxels. Unlike Voxformer \cite{li2023voxformer}, which learns class-agnostic proposal on temporal data \textbf{offline} for voxel queries, our SVPN aims to dynamically select seed voxels \textbf{online} by occupancy probability for subsequent semantic context learning. Specifically, SVPN consists of depth estimation and coarse-to-fine occupancy prediction. Following \cite{li2023voxformer, jiang2024symphonize}, we utilize the pre-trained Mobilestereonet \cite{shamsafar2022mobilestereonet} to infer the depth prediction and calculate the scene points $\textbf{P}$ by back-projecting the depth map into the 3D point cloud using the camera parameters $(\textbf{K}, \textbf{T})$. The scene points $\textbf{P}$ imply the volumetric surface and are used for occupancy prediction. Let $(u, v)$ denote a pixel in the depth map and $p$ is the corresponding 3D point, the back-projecting procedure is formulated as follows:
\begin{equation}
    p = \textbf{R}^{-1}\cdot [\textbf{K}^{-1}\cdot (z_c \circ [u, v, 1]^T) - \textbf{t}]
\end{equation} where $\circ$ denote element-wise product, $z_c$ is the depth value of the pixel. $\textbf{K}$ and $\textbf{T}=[\textbf{R}, \textbf{t}]$ are the intrinsic and extrinsic parameters of the camera.

Next, we generate the occupancy prediction $\textbf{O} \in \mathbb{R}^{X \times Y \times Z}$ in a coarse-to-fine manner. Firstly, the points $\textbf{P}$ are fed into a voxelization layer adopted from DRNet \cite{ye2021drinet} for voxel-wise features. Then we apply the tiny sparse convolution network consisting of a sparse feature encoder and a sparse geometry feature encoder adopted from GASN \cite{ye2022efficient} to predict the coarse occupancy probability from the voxel-wise features. Finally, the occupancy probability is further fed into a lightweight Unet-like network \cite{roldao2020lmscnet} for the final occupancy prediction $\textbf{O}$, which is used to select the sparse voxels as explained in semantic guidance. 
Similar to the geometry guidance, we use the binary cross entropy loss to calculate the loss $\mathcal{L}_{occ}$ for occupancy prediction.

To further utilize the geometry information from the depth-based SVPN, we also take the occupancy-aware 3D features $\textbf{F}_{o}^{3D} \in \mathbb{R}^{C_o \times X \times Y \times Z} $ from the last layer of the Unet-like network for the latter voxel feature aggregation.

\paragraph{Semantic Guidance} After obtaining the occupancy prediction $\textbf{O}$ and voxel coordinates $\textbf{V}^{3D} \in \mathbb{Z}^{3 \times X \times Y \times Z}$ of the scene, we first choose the initial seed voxel features $\textbf{F}_{s, 0}^{3D} \in \mathbb{R}^{C \times N_s}$ and seed coordinates $\textbf{V}_{s}^{3D} \in \mathbb{Z}^{3 \times N_s}$ by:
\begin{equation}
    \textbf{V}_{s}^{3D} = \textbf{V}^{3D}[:, \textbf{O} > \theta]
\end{equation}
\begin{equation}
    \textbf{F}_{s, 0}^{3D} = \textbf{F}^{3D}[:, \textbf{O} > \theta]
\end{equation} where $\theta$ is the threshold to determine if the voxel is occupied and $N_s$ is the number of non-empty voxel. 
Then these seed voxel features $\textbf{F}_{s, 0}^{3D}$ and corresponding voxel indices $V_{s}^{3D}$ are fed into the semantic guidance module (SGM) illustrated in Figure \ref{fig:sdb} for mutual interactions. The semantic guidance module has two sparse encoder blocks (SEB), a fusion layer, and an auxiliary semantic head. Each sparse encoder block (SEB) consists of a sparse feature encoder and a sparse geometry feature encoder adopted from \cite{ye2022efficient} and outputs features with multi-scale contextual information. Let $\textbf{F}_{s,1}^{3D}, \textbf{F}_{s,2}^{3D}$ are the outputs of the two sparse encoder blocks, the fusion feature $\textbf{F}_{s}^{3D} \in \mathbb{R}^{C \times N_s}$ are obtained by:
\begin{equation}
    \textbf{F}_{s}^{3D} = \mathrm{MLP}([\textbf{F}_{s, 0}^{3D}, \textbf{F}_{s,1}^{3D}, \textbf{F}_{s,2}^{3D}])
\end{equation} where $[.]$ denotes concatenate operation along feature dimension. After that, the fused features $\textbf{F}_{s}^{3D}$ are fed into the auxiliary semantic head consisting of a two-layer MLP to predict the corresponding semantics $\hat{\textbf{Y}}_{s} \in \mathbb{R}^{C_{class} \times N_s}$, where $C_{class}$ is the number of classes. We calculate the cross entropy loss and lovasz loss \cite{berman2018lovasz} for the semantic guidance:
\begin{equation}
    \mathcal{L}_{sem} = \mathcal{L}_{ce}(\hat{\textbf{Y}}_{s}, \textbf{Y}_s) + \mathcal{L}_{lovasz}(\hat{\textbf{Y}}_{s}, \textbf{Y}_s)
\end{equation} where $\textbf{Y}_s$ is the seed voxels' semantic label indexed from the semantic scene label $\textbf{Y}$.

By this means, we inject semantic cues into the fused seed features $\textbf{F}_{s}^{3D}$ and enhance the feature separation between categories, which is the key to semantic propagation.

\newcommand{\note}[1]{\textbf{#1}}

\definecolor{White}{rgb}{1.,0.,1.}
\definecolor{first}{rgb}{.8,.0,.0}
\definecolor{second}{rgb}{.0,.6,.0}
\definecolor{third}{rgb}{.0,.0,.8}
\newcolumntype{g}{>{\columncolor{White}}c}

\definecolor{car}{rgb}{0.39215686, 0.58823529, 0.96078431}
\definecolor{bicycle}{rgb}{0.39215686, 0.90196078, 0.96078431}
\definecolor{motorcycle}{rgb}{0.11764706, 0.23529412, 0.58823529}
\definecolor{truck}{rgb}{0.31372549, 0.11764706, 0.70588235}
\definecolor{othervehicle}{rgb}{0.39215686, 0.31372549, 0.98039216}
\definecolor{person}{rgb}{1.        , 0.11764706, 0.11764706}
\definecolor{bicyclist}{rgb}{1.        , 0.15686275, 0.78431373}
\definecolor{motorcyclist}{rgb}{0.58823529, 0.11764706, 0.35294118}
\definecolor{road}{rgb}{1.        , 0.        , 1.        }
\definecolor{parking}{rgb}{1.        , 0.58823529, 1.        }
\definecolor{sidewalk}{rgb}{0.29411765, 0.        , 0.29411765}
\definecolor{otherground}{rgb}{0.68627451, 0.        , 0.29411765}
\definecolor{building}{rgb}{1.        , 0.78431373, 0.        }
\definecolor{fence}{rgb}{1.        , 0.47058824, 0.19607843}
\definecolor{vegetation}{rgb}{0.        , 0.68627451, 0.        }
\definecolor{trunk}{rgb}{0.52941176, 0.23529412, 0.        }
\definecolor{terrain}{rgb}{0.58823529, 0.94117647, 0.31372549}
\definecolor{pole}{rgb}{1.        , 0.94117647, 0.58823529}
\definecolor{trafficsign}{rgb}{1.        , 0.        , 0.    }
\definecolor{otherstructure}{rgb}{0.98039215, 0.58823529, 0.}
\definecolor{otherobject}{rgb}{0.19607843, 1.        , 1.        }
\newcommand*\rot{\rotatebox{90}}

\makeatletter
\newcommand{\car@semkitfreq}{3.92}
\newcommand{\bicycle@semkitfreq}{0.03}
\newcommand{\motorcycle@semkitfreq}{0.03}
\newcommand{\truck@semkitfreq}{0.16}
\newcommand{\othervehicle@semkitfreq}{0.20}
\newcommand{\person@semkitfreq}{0.07}
\newcommand{\bicyclist@semkitfreq}{0.07}
\newcommand{\motorcyclist@semkitfreq}{0.05}
\newcommand{\road@semkitfreq}{15.30}  %
\newcommand{\parking@semkitfreq}{1.12}
\newcommand{\sidewalk@semkitfreq}{11.13}  %
\newcommand{\otherground@semkitfreq}{0.56}
\newcommand{\building@semkitfreq}{14.1}  %
\newcommand{\fence@semkitfreq}{3.90}
\newcommand{\vegetation@semkitfreq}{39.3}  %
\newcommand{\trunk@semkitfreq}{0.51}
\newcommand{\terrain@semkitfreq}{9.17} %
\newcommand{\pole@semkitfreq}{0.29}
\newcommand{\trafficsign@semkitfreq}{0.08}
\newcommand{\semkitfreq}[1]{{\csname #1@semkitfreq\endcsname}}

\newcommand{\car@sscbkitfreq}{2.85}
\newcommand{\bicycle@sscbkitfreq}{0.01}
\newcommand{\motorcycle@sscbkitfreq}{0.01}
\newcommand{\truck@sscbkitfreq}{0.16}
\newcommand{\othervehicle@sscbkitfreq}{5.75}
\newcommand{\person@sscbkitfreq}{0.02}
\newcommand{\road@sscbkitfreq}{14.98}
\newcommand{\parking@sscbkitfreq}{2.31}
\newcommand{\sidewalk@sscbkitfreq}{6.43}
\newcommand{\otherground@sscbkitfreq}{2.05}
\newcommand{\building@sscbkitfreq}{15.67}
\newcommand{\fence@sscbkitfreq}{0.96}
\newcommand{\vegetation@sscbkitfreq}{41.99}
\newcommand{\terrain@sscbkitfreq}{7.10}
\newcommand{\pole@sscbkitfreq}{0.22}
\newcommand{\trafficsign@sscbkitfreq}{0.06}
\newcommand{\otherstructure@sscbkitfreq}{4.33}
\newcommand{\otherobject@sscbkitfreq}{0.28}
\newcommand{\sscbkitfreq}[1]{{\csname #1@sscbkitfreq\endcsname}}

\begin{table*}[t]
    \small
    \newcommand{\classfreq}[1]{{~\tiny(\semkitfreq{#1}\%)}}  %
    \newcommand{\clsname}[1]{\rotatebox{90}{\textcolor{#1}{$\blacksquare$} #1\classfreq{#1}}}
    \renewcommand{\tabcolsep}{2pt}
    \renewcommand\arraystretch{1.25}
    \centering
    \caption{\textbf{Semantic Scene Completion on SemanticKITTI hidden \texttt{test} set.} $^\dagger$ denotes the results provided by MonoScene \cite{cao2022monoscene}. Bold and underline denote the best performance and the second-best performance, respectively.}
    \scalebox{0.95}{
    \begin{tabular}{l|c|ccccccccccccccccccc|c}
    \toprule
        Method & IoU 
        & \clsname{road}
        & \clsname{sidewalk}
        & \clsname{parking}
        & \clsname{otherground}
        & \clsname{building}
        & \clsname{car}
        & \clsname{truck}
        & \clsname{bicycle}
        & \clsname{motorcycle}
        & \clsname{othervehicle}
        & \clsname{vegetation}
        & \clsname{trunk}
        & \clsname{terrain}
        & \clsname{person}
        & \clsname{bicyclist}
        & \clsname{motorcyclist}
        & \clsname{fence}
        & \clsname{pole}
        & \clsname{trafficsign}
        & mIoU
        \\
    \midrule
        LMSCNet$^\dagger$ \cite{roldao2020lmscnet} & 31.38 & 46.70 & 19.50 & 13.50 & 3.10 & 10.30 & 14.30 & 0.30 & 0.00 & 0.00 & 0.00 & 10.80 & 0.00 & 10.40 & 0.00 & 0.00 & 0.00 & 5.40 & 0.00 & 0.00 & 7.07 \\
        AICNet$^\dagger$ \cite{li2020anisotropic} & 23.93 & 39.30 & 18.30 & 19.80 & 1.60 & 9.60 & 15.30 & 0.70 & 0.00 & 0.00 & 0.00 & 9.60 & 1.90 & 13.50 & 0.00 & 0.00 & 0.00 & 5.00 & 0.10 & 0.00 & 7.09 \\
        JS3C-Net$^\dagger$ \cite{yan2021sparse} & 34.00 & 47.30 & 21.70 & 19.90 & 2.80 & 12.70 & 20.10 & 0.80 & 0.00 & 0.00 & 4.10 & 14.20 & 3.10 & 12.40 & 0.00 & 0.20 & 0.20 & 8.70 & 1.90 & 0.30 & 8.97 \\
        MonoScene \cite{cao2022monoscene} & 34.16 & 54.70 & 27.10 & 24.80 & 5.70 & 14.40 & 18.80 & 3.30 & 0.50 & 0.70 & \underline{4.40} & 14.90 & 2.40 & 19.50 & 1.00 & 1.40 & \underline{0.40} & 11.10 & 3.30 & 2.10 & 11.08 \\
        TPVFormer \cite{huang2023tri} & 34.25 & 55.10 & 27.20 & 27.40 & 6.50 & 14.80 & 19.20 & 3.70 & 1.00 & 0.50 & 2.30 & 13.90 & 2.60 & 20.40 & 1.10 & 2.40 & 0.30 & 11.00 & 2.90 & 1.50 & 11.26 \\
        VoxFormer \cite{li2023voxformer} & 42.95 & 53.90 & 25.30 & 21.10 & 5.60 & 19.80 & 20.80 & 3.50 & 1.00 & 0.70 & 3.70 & 22.40 & 7.50 & 21.30 & 1.40 & \underline{2.60} & 0.20 & 11.10 & 5.10 & 4.90 & 12.20 \\
        OccFormer \cite{zhang2023occformer} & 34.53 & 55.90 & \underline{30.30} & \textbf{31.50} & 6.50 & 15.70 & 21.60 & 1.20 & 1.50 & \underline{1.70} & 3.20 & 16.80 & 3.90 & 21.30 & \textbf{2.20} & 1.10 & 0.20 & 11.90 & 3.80 & 3.70 & 12.32 \\
        SurroundOcc \cite{wei2023surroundocc} & 34.72 & 56.90 & 28.30 & \underline{30.20} & \underline{6.80} & 15.20 & 20.60 & 1.40 & \underline{1.60} & 1.20 & \underline{4.40} & 14.90 & 3.40 & 19.30 & 1.40 & 2.00 & 0.10 & 11.30 & 3.90 & 2.40 & 11.86 \\
        NDC-scene \cite{yao2023ndc} & 36.19 & 58.12 & 28.05 &25.31& 6.53& 14.90& 19.13& \textbf{4.77}& \textbf{1.93}& \textbf{2.07}& \textbf{6.69}& 17.94& 3.49& 25.01& \textbf{3.44} &\textbf{2.77}& \textbf{1.64}& 12.85& 4.43& 2.96& 12.58 \\
        \hline
        \textbf{SGN-S} (ours) & 41.88 & 57.80 & 29.20 & 27.70 & 5.20 & 23.90 & 24.90 & 2.70 & 0.40 & 0.30 & 4.00 & 24.20 & 10.00 & 25.80 & 1.10 & 2.50 & 0.30 & 14.20 & 7.40 & 4.40 & 14.01\\
        \textbf{SGN-L} (ours) & \underline{43.71} & \underline{57.90} & 29.70 & 25.60 & 5.50 & \underline{27.00} & \underline{25.00} & 1.50 & 0.90 & 0.70 & 3.60 & \underline{26.90} & \underline{12.00} & \underline{26.40} & 0.60 & 0.30 & 0.00 & \underline{14.70} & \underline{9.00} & \underline{6.40} & \underline{14.39} \\
        \textbf{SGN-T} (ours) & \textbf{45.42} & \textbf{60.40} & \textbf{31.40} & 28.90 & \textbf{8.70} & \textbf{28.40} & \textbf{25.40} & \underline{4.50} & 0.90 & 1.60 & 3.70 & \textbf{27.40} & \textbf{12.60} & \textbf{28.40} & 0.50 & 0.30 & 0.10 & \textbf{18.10} & \textbf{10.00} & \textbf{8.30} & \textbf{15.76}\\
    \bottomrule
    \end{tabular}}
    \label{tab:semkitti_test}
\end{table*}

\begin{table*}[ht]
    \small
    \newcommand{\classfreq}[1]{{~\tiny(\semkitfreq{#1}\%)}}  %
    \newcommand{\clsname}[1]{\rotatebox{90}{\textcolor{#1}{$\blacksquare$} #1\classfreq{#1}}}
    \renewcommand{\tabcolsep}{2pt}
    \renewcommand\arraystretch{1.25}
    \centering
    \caption{\textbf{Semantic Scene Completion on SemanticKITTI \texttt{val} set.} $^\dagger$ denotes the results provided by MonoScene. Bold and underline denote the best performance and the second-best performance, respectively.}
    \scalebox{0.95}{
    \begin{tabular}{l|c|ccccccccccccccccccc|c}
    \toprule
        Method & IoU
        & \clsname{road}
        & \clsname{sidewalk}
        & \clsname{parking}
        & \clsname{otherground}
        & \clsname{building}
        & \clsname{car}
        & \clsname{truck}
        & \clsname{bicycle}
        & \clsname{motorcycle}
        & \clsname{othervehicle}
        & \clsname{vegetation}
        & \clsname{trunk}
        & \clsname{terrain}
        & \clsname{person}
        & \clsname{bicyclist}
        & \clsname{motorcyclist}
        & \clsname{fence}
        & \clsname{pole}
        & \clsname{trafficsign}
        & mIoU
        \\
    \midrule
        LMSCNet$^\dagger$ \cite{roldao2020lmscnet} & 28.61 & 40.68 & 18.22 & 4.38 & 0.00 & 10.31 & 18.33 & 0.00 & 0.00 & 0.00 & 0.00 & 13.66 & 0.02 & 20.54 & 0.00 & 0.00 & 0.00 & 1.21 & 0.00 & 0.00 & 6.70 \\
        AICNet$^\dagger$ \cite{li2020anisotropic} & 29.59 & 43.55 & 20.55 & 11.97 & 0.07 & 12.94 & 14.71 & 4.53 & 0.00 & 0.00 & 0.00 & 15.37 & 2.90 & 28.71 & 0.00 & 0.00 & 0.00 & 2.52 & 0.06 & 0.00 & 8.31 \\
        JS3C-Net$^\dagger$ \cite{yan2021sparse} & 38.98 & 50.49 & 23.74 & 11.94 & 0.07 & 15.03 & 24.65 & 4.41 & 0.00 & 0.00 & 6.15 & 18.11 & 4.33 & 26.86 & 0.67 & 0.27 & 0.00 & 3.94 & 3.77 & 1.45 & 10.31 \\
        MonoScene \cite{cao2022monoscene} & 37.12 & 57.47 & 27.05 & 15.72 & \underline{0.87} & 14.24 & 23.55 & 7.83 & 0.20 & 0.77 & 3.59 & 18.12 & 2.57 & 30.76 & 1.79 & 1.03 & 0.00 & 6.39 & 4.11 & 2.48 & 11.50 \\
        TPVFormer \cite{huang2023tri} & 35.61 & 56.50 & 25.87 & \underline{20.60} & 0.85 & 13.88 & 23.81 & 8.08 & 0.36 & 0.05 & 4.35 & 16.92 & 2.26 & 30.38 & 0.51 & 0.89 & 0.00 & 5.94 & 3.14 & 1.52 & 11.36 \\
        VoxFormer \cite{li2023voxformer} & 44.02 & 54.76 & 26.35 & 15.50 & 0.70 & 17.65 & 25.79 & 5.63 & 0.59 & 0.51 & 3.77 & 24.39 & 5.08 & 29.96 & 1.78 & \textbf{3.32} & 0.00 & 7.64 & 7.11 & 4.18 & 12.35 \\
        OccFormer \cite{zhang2023occformer} & 36.50 & 58.85 & 26.88 & 19.61 & 0.31 & 14.40 & 25.09 & \textbf{25.53} & \underline{0.81} & \underline{1.19} & \underline{8.52} & 19.63 & 3.93 & 32.62 & \underline{2.78} & \underline{2.82} & 0.00 & 5.61 & 4.26 & 2.86 & 13.46 \\ 
        NDC-scene \cite{yao2023ndc}&37.24 &59.20& 28.24& \textbf{21.42}& \textbf{1.67} &14.94& 26.26 &\underline{14.75} &\textbf{1.67}& \textbf{2.37}& 7.73& 19.09& 3.51& 31.04 &\textbf{3.60}& 2.74& 0.00& 6.65& 4.53& 2.73& 12.70\\
        \hline
        \textbf{SGN-S} (ours) & 43.60 & \textbf{59.32} & \textbf{30.51} & 18.46 & 0.42 & 21.43 & 31.88 & 13.18 & 0.58 & 0.17 & 5.68 & 25.98 & 7.43 & 34.42 & 1.28 & 1.49 & 0.00 & \underline{9.66} & 9.83 & 4.71 & 14.55 \\ 
        \textbf{SGN-L} (ours) & \underline{45.45} & 59.00 & \underline{30.11} & 19.35 & 0.21 & \underline{23.95} & \underline{32.51} & 9.74 & 0.39 & 0.15 & 5.19 & \underline{28.29} & \underline{8.48} & \underline{34.91} & 0.78 & 0.20 & 0.00 & 8.83 & \underline{12.13} & \underline{6.95} & \underline{14.80} \\
        \textbf{SGN-T} (ours) & \textbf{46.21} & \underline{59.10} & 29.41 & 19.05 & 0.33 & \textbf{25.17} & \textbf{33.31} & 6.03 & 0.61 & 0.46 & \textbf{9.84} & \textbf{28.93} & \textbf{9.58} & \textbf{38.12} & 0.47 & 0.10 & 0.00 & \textbf{9.96} & \textbf{13.25} & \textbf{7.32} & \textbf{15.32} \\
    \bottomrule
    \end{tabular}}
    \label{tab:semkitti_val}
\end{table*}

\begin{table}[t]\centering
    \small
    \renewcommand\tabcolsep{1.5pt}
    \renewcommand\arraystretch{1.1}
    \caption{\textbf{Quantitative comparison} in different ranges on SemanticKITTI validation. ``*" denotes the results provided by VoxFormer.}
    \scalebox{0.95}{
    \begin{tabular}{l|c|ccc|ccc}
    \toprule
        \multirow{2}{*}{Methods}
        & \multirow{2}{*}{Modality} 
        & \multicolumn{3}{c|}{IoU (\%)} 
        & \multicolumn{3}{c}{mIoU (\%)} \\
        &  & 12.8m & 25.6m & 51.2m & 12.8m & 25.6m & 51.2m \\
    \midrule
        SSCNet \cite{song2017semantic} & LiDAR & 64.37 & 61.02 & \underline{50.22} &20.02 & 19.68 & \underline{16.35} \\
        JS3CNet \cite{yan2021sparse} & LiDAR &63.47 & \textbf{63.40} & \textbf{53.09} & \textbf{30.55} & \textbf{28.12} & \textbf{22.67} \\
    \midrule
        MonoScene* \cite{cao2022monoscene} & Camera & 38.42 & 38.55 & 36.80 & 12.25 & 12.22 & 11.30 \\
        OccFormer \cite{zhang2023occformer} & Camera & 56.38 &47.28 &36.50 & 20.91 &17.90 &13.46 \\
        VoxFormer-S \cite{li2023voxformer} & Camera & 65.35 & 57.54 & 44.02 & 17.66 & 16.48 & 12.35 \\
        VoxFormer-T \cite{li2023voxformer} & Camera & 65.38 &57.69 &44.15 & 21.55 &18.42 &13.35 \\
        \hline
        \textbf{SGN-S} (ours) & Camera & 64.21 & 56.20 & 43.60 & 21.53 & 19.60 & 14.55\\
        \textbf{SGN-L} (ours) & Camera & \underline{70.08} & 61.17 & 45.45 & 24.76 & 21.17 & 14.80\\
        \textbf{SGN-T} (ours) & Camera & \textbf{70.61} & \underline{61.90} & 46.21 & \underline{25.70} & \underline{22.02} & 15.32 \\
    \bottomrule
    \end{tabular}}
    \label{tab:comp2lidar}
\end{table}

\begin{table*}[ht]
    \centering
    \caption{\textbf{Quantitative results on SSCBench-KITTI360 test set.} The results for counterparts are provided in \cite{li2023sscbench}. Bold and underline denote the best performance and the second-best performance, respectively.}
    \newcommand{\clsname}[2]{
        \rotatebox{90}{
            \hspace{-6pt}
            \textcolor{#2}{$\blacksquare$}
            \hspace{-6pt}
            \renewcommand\arraystretch{0.6}
            \begin{tabular}{l}
                #1~\tiny(\sscbkitfreq{#2}\%) \\
            \end{tabular}
        }}
    \renewcommand{\tabcolsep}{2pt}
    \renewcommand\arraystretch{1.2}
    \scalebox{0.95}{ 
        \begin{tabular}{l|cccc|cccccccccccccccccc}
            \toprule
            \multicolumn{1}{c|}{Method} &\multicolumn{1}{c}{IoU}        &
            \multicolumn{1}{c}{Precision}                               &
            \multicolumn{1}{c}{Recall}                                  &
            mIoU                                                        &
            \multicolumn{1}{c}{\clsname{car}{car}}                      &
            \multicolumn{1}{c}{\clsname{bicycle}{bicycle}}              &
            \multicolumn{1}{c}{\clsname{motorcycle}{motorcycle}}        &
            \multicolumn{1}{c}{\clsname{truck}{truck}}                  &
            \multicolumn{1}{c}{\clsname{other-veh.}{othervehicle}}      &
            \multicolumn{1}{c}{\clsname{person}{person}}                &
            \multicolumn{1}{c}{\clsname{road}{road}}                    &
            \multicolumn{1}{c}{\clsname{parking}{parking}}              &
            \multicolumn{1}{c}{\clsname{sidewalk}{sidewalk}}            &
            \multicolumn{1}{c}{\clsname{other-grnd.}{otherground}}      &
            \multicolumn{1}{c}{\clsname{building}{building}}            &
            \multicolumn{1}{c}{\clsname{fence}{fence}}                  &
            \multicolumn{1}{c}{\clsname{vegetation}{vegetation}}        &
            \multicolumn{1}{c}{\clsname{terrain}{terrain}}              &
            \multicolumn{1}{c}{\clsname{pole}{pole}}                    &
            \multicolumn{1}{c}{\clsname{traf.-sign}{trafficsign}}       &
            \multicolumn{1}{c}{\clsname{other-struct.}{otherstructure}} &
            \multicolumn{1}{c}{\clsname{other-obj.}{otherobject}}
            \\
            \midrule
            \multicolumn{23}{l}{\textit{LiDAR-based}}\\
            \hline
            SSCNet \cite{song2017semantic} & \textbf{53.58} & \underline{69.63} & \textbf{69.92} & {16.95} & \textbf{31.95} & 0.00 & 0.17 & 10.29 & 0.00 & 0.07 & \textbf{65.70} & \textbf{17.33} & \textbf{41.24} & 3.22 & \textbf{44.41} & 6.77 & \textbf{43.72} & \textbf{28.87} & 0.78 & 0.75 & 8.69 & 0.67\\
            LMSCNet \cite{roldao2020lmscnet} & \underline{47.35} & \textbf{72.77} & 57.55 & 13.65 & 20.91 & 0.00 & 0.00 & 0.26 & 0.58 & 0.00 & \underline{62.95} & 13.51 & 33.51 & 0.20 & \underline{43.67} & 0.33 & \underline{40.01} & \underline{26.80} & 0.00 & 0.00 & 3.63 & 0.00 \\
            \hline
            \multicolumn{23}{l}{\textit{Camera-based}}\\
            \hline
            MonoScene \cite{cao2022monoscene} & 37.87 & 56.73 & 53.26 & 12.31 & 19.34 & 0.43 & 0.58 & 8.02 & 2.03 & 0.86 & 48.35 & 11.38 & 28.13 & 3.32 & 32.89 & 3.53 & 26.15 & 16.75 & 6.92 & 5.67 & 4.20 & 3.09 \\
            TPVFormer \cite{huang2023tri} & 40.22 & 59.32 & 55.54 & 13.64        & 21.56 & 1.09 & 1.37 & 8.06 & 2.57 & 2.38 & 52.99 & 11.99 & 31.07 & 3.78 & 34.83 & 4.80 & 30.08 & 17.52 & 7.46 & 5.86 & 5.48 & 2.70 \\
            VoxFormer \cite{li2023voxformer} & 38.76 & 58.52 & 53.44 & 11.91 & 17.84 & 1.16 & 0.89 & 4.56 & 2.06 & 1.63 & 47.01 & 9.67 & 27.21 & 2.89 & 31.18 & 4.97 & 28.99 & 14.69 & 6.51 & 6.92 & 3.79 & 2.43 \\
            OccFormer \cite{zhang2023occformer} & 40.27 & 59.70 & 55.31 & 13.81 & 22.58 & 0.66 & 0.26 & 9.89 & 3.82 & 2.77 & 54.30 & 13.44 & 31.53 & 3.55 & 36.42 & 4.80 & 31.00 & 19.51 & 7.77 & 8.51 & 6.95 & 4.60 \\
            DepthSSC \cite{yao2023depthssc} & 40.85 & 60.69 & 55.86 & 14.28 & 21.90 & \underline{2.36} & 4.30 & 11.51 & 4.56 & 2.92 & 50.88 & 12.89 & 30.27 & 2.49 & 37.33 & 5.22 & 29.61 & 21.59 & 5.97 & 7.71 & 5.24 & 3.51 \\
            Symphonize \cite{jiang2024symphonize} & 44.12 & 69.24 & 54.88 & \textbf{18.58} & \underline{30.02} & 1.85 & 5.90 & \textbf{25.07} & \textbf{12.06} & \textbf{8.20} & 54.94 & 13.83 & 32.76 & \textbf{6.93} & 35.11 & \textbf{8.58} & 38.33 & 11.52 & 14.01 & 9.57 & \textbf{14.44} & \textbf{11.28} \\
            \hline
            SGN-S (ours) & 46.22 & 68.17 & 58.94 & 17.71 & 28.20 & {2.09} & {3.02} & \underline{11.95} & 3.68 & \underline{4.20} & 59.49 & 14.50 & \underline{36.53} & {4.24} & 39.79 & {7.14} & 36.61 & 23.10 & 14.86 & 16.14 & {8.24} & 4.95  \\
            SGN-L (ours) & 46.64 & 68.26 &59.55 &16.95 & 26.91 & 1.72 &0.85 & 8.60 & {3.80} & 1.93 & 56.52 & 13.83 & 35.40 & 3.42 & 40.62 & 6.65 & 36.68 & 22.00 & \underline{15.84} & \textbf{16.49} & 8.06 & {5.76}  \\
            SGN-T (ours) & 47.06 & 68.83 & \underline{59.81} & \underline{18.25} & {29.03} & \textbf{3.43} & {2.90} & {10.89} & \underline{5.20} & {2.99} & 58.14 & \underline{15.04} & 36.40 & \underline{4.43} & 42.02 & \underline{7.72} & 38.17 & 23.22 & \textbf{16.73} & \underline{16.38} & \underline{9.93} & \underline{5.86}  \\
            \bottomrule
        \end{tabular}
    }
    \label{tab:kitti_360_test}
\end{table*}

\subsection{Voxel Aggregation} As shown in Figure \ref{fig:pipeine}, to fully exploit the geometry information in 3D features $\textbf{F}^{3D}$ and $\textbf{F}_{o}^{3D}$,  we further aggregate them with the semantic-aware seed features $\textbf{F}_{s}^{3D}$ to construct the final discriminative voxel features $\textbf{F}_f^{3D} \in \mathbb{R}^{(C+C_o) \times X \times Y \times Z}$ for subsequent semantic propagation. Specifically, we leverage the coordinates of non-seed voxels to index features $\textbf{F}_{n}^{3D}$ from $\textbf{F}^{3D}$. Then the non-seed voxel features $\textbf{F}_{n}^{3D}$ are fed into a linear layer and combined with the semantic-aware features $\textbf{F}_{s}^{3D}$ to form the new scene representation, which contains the semantic context and geometry cues. We argue that non-seed voxel features $\textbf{F}_{n}^{3D}$ are vital and can well complement the seed features since the sparse voxel proposal is not always accurate. To further utilize the geometry information from the SVPN, we also concatenate the features $\textbf{F}_{o}^{3D}$ from SVPN with the new scene representation to obtain the final voxel features. The detailed procedure can be formulated as follows:
\begin{equation}
    \textbf{F}_f^{3D} = \mathrm{MLP}([\mathrm{CN}(\textbf{F}_{s}^{3D}, \mathrm{Conv1d}(\textbf{F}_{n}^{3D})), \textbf{F}_{o}^{3D}])
\end{equation} where $\mathrm{CN}$ is the feature combination of seed and non-seed voxels.
\subsection{Multi-Scale Semantic Propagation}
By learning features with hybrid guidance and voxel aggregation, we obtain discriminative voxel features $\textbf{F}_f^{3D}$ with the rich semantic context in the seed features $\textbf{F}_{s}^{3D}$ and spatial geometry cues from previous 3D features $\textbf{F}_{n}^{3D}$, and occupancy-aware features $\textbf{F}_{o}^{3D}$. Then we design the multi-scale semantic propagation (MSSP) module to propagate the semantic information from seed features to the whole scene. The MSSP module contains three anisotropic convolutional layers \cite{li2020anisotropic} and the ASPP \cite{chen2017deeplab} module consisting of three 3D convolutions with the kernel size of $3\times3\times3$ and dilation of 1, 2, and 4. This module is lightweight and can well capture multi-scale features of instances of different sizes through convolutional kernels with different receptive fields. Afterward, we use the head consisting of a linear layer and softmax layer to predict the final semantic scene prediction $\hat{\textbf{Y}} \in \mathbb{R}^{C_{class} \times X \times Y \times Z}$ from the propagated voxel features. 

Following MonoScene \cite{cao2022monoscene}, we adopt the Scene-Class Affinity Loss to force the network to account for voxels within the same category as well as voxels across different categories. The Affinity Loss optimizes the class-wise derivable precision, recall, and specificity metrics simultaneously, where precision and recall evaluate the performance of voxels within the same class, while specificity assesses the performance of dissimilar voxels. Specifically, similar to \cite{cao2022monoscene, yao2023ndc}, we apply scene-class affinity loss on both semantic and geometry results of the prediction $\hat{\textbf{Y}}$. We integrate and optimize scene- and class-wise semantics $\mathcal{L}_{scal}^{sem}$, geometry $\mathcal{L}_{scal}^{geo}$, and cross-entropy loss $\mathcal{L}_{ce}$. The overall loss function is formulated by:
\begin{equation}
    \mathcal{L}_{ssc} = \mathcal{L}_{scal}^{sem}(\hat{\textbf{Y}}, \textbf{Y}) + \mathcal{L}_{scal}^{geo}(\hat{\textbf{Y}}, \textbf{Y}) + \mathcal{L}_{ce}(\hat{\textbf{Y}}, \textbf{Y})
\end{equation}

\subsection{Training Loss}
Unlike VoxFormer \cite{li2023voxformer} with sophisticated two-stage training, we train our SGN end-to-end. The total training loss $\mathcal{L} = \mathcal{L}_{geo} + \mathcal{L}_{occ} + \mathcal{L}_{sem} + \mathcal{L}_{ssc}$.

\section{Experiments}
In this section, we present the datasets, evaluation metrics, and detailed implementation aspects of our approach. Subsequently, we conduct extensive experiments to establish that our proposed SGN consistently surpasses or achieves comparable performance against the state-of-the-art methods on the complex, large-scale outdoor dataset SemanticKITTI \cite{behley2019semantickitti} as well as SSCBench-KITTI-360 \cite{li2023sscbench}. Following this, we provide qualitative results to underscore the efficacy of our SGN. Moreover, we conducted detailed ablation studies to dissect the contribution of individual components of our method and various configurations, thereby offering an in-depth analysis of our approach. Additionally, we provide detailed experiments on the NYUv2 dataset \cite{silberman2012indoor} to demonstrate the generalization ability of our SGN on indoor scenes.

\subsection{Dataset and Metrics}
\paragraph{Dataset} For large-scale outdoor scene understanding, the KITTI odometry dataset \cite{geiger2012we} collects 22 sequences with 20 classes with a Velodyne HDL-64 laser scanner in the scenes of autonomous driving. SemanticKITTI \cite{behley2019semantickitti} is based on the KITTI dataset and provides semantic annotation of all sequences. According to the official setting for semantic scene completion (SSC), sequences 00-07 and 09-10 (a total of 3834 scans) are for training, sequence 08 (815 scans) is for validation, and the rest (3901 scans) is for testing. 
SSCBench-KITTI-360 \cite{li2023sscbench} offers a comprehensive benchmark for semantic scene completion, featuring nine densely annotated sequences of urban driving scenes. The dataset is meticulously partitioned, with the training set encompassing 8,487 frames across scenes 00, 02-05, 07, and 10. The validation set is carefully curated with 1,812 frames from scene 06, ensuring a robust evaluation framework. Furthermore, the testing set includes 2,566 frames exclusively from scene 09, providing a diverse and challenging environment for model assessment.
The volume of interest for both two SSC benchmarks is $[0 \sim 51.2m, -25.6m \sim 25.6m, -2 \sim 4.4m]$, and the voxelization resolution s is $0.2m$. The SSC labels with resolution $256\times256\times32$ of train and validation set are provided for the users. 
In this work, we focus on the camera-based SSC, taking the RGB images as inputs similar to \cite{cao2022monoscene, li2023voxformer, jiang2024symphonize}.

\paragraph{Metrics} Following \cite{song2017semantic}, we mainly report the Intersection-over-Union (IoU) for scene completion and mIoU of $C_n$ classes (no ``unlabeled" class) for semantic scene completion. The mIoU is calculated by:
\begin{equation}
    mIoU = \frac{1}{C_n}\sum_{c=1}^{C_n}{\frac{TP_c}{TN_c+FP_c+FN_c}}
\end{equation} where $TP_c$, $TN_c$, $FP_c$, and $FN_c$ denote true positive, true negative, false positive, and false negative for class $c$.

\subsection{Implementation Details}
We crop the input RGB images of cam2 to size $1220\times 370$ for SemanticKITTI and RGB images of cam1 of $1408\times 376$ for SSCBench-KITTI-360. The 2D feature maps with 1/16 of the input resolution are taken for the subsequent processing. The feature dimension $C$ and the channel number $C_o$ are set to 128 and 8, respectively. The size $X \times Y \times Z$ of the 3D feature volume is $128\times 128 \times 16$. And the final predictions are up-sampled to $256\times 256 \times 32$. The threshold $\theta$ for selecting seed voxels is set to 0.5 by default.
We train SGN for 40 epochs on 4 V100 GPUs with a total batch size of 4. The AdamW \cite{loshchilov2017decoupled} optimizer is used with an initial learning rate of 2e-4 and a weight decay of 1e-2.
Following VoxFormer \cite{li2023voxformer}, we design the single-image version SGN-S, taking only the current image as input and the temporal version SGN-T with the current and the previous 4 images as input. We also provide the lightweight version SGN-L, which takes temporal inputs and uses ResNet18 as the backbone with dimension $C=64$ and 1 anisotropic convolution layer for MSSP.

\begin{table}[t]\centering
    \small
    \renewcommand\tabcolsep{4.9pt}
    \renewcommand\arraystretch{1.1}
    \caption{Ablation on network components, i.e., semantic guidance (SG), geometry guidance (GG), multi-scale semantic
propagation (MSSP), and voxel aggregation layer (VA).}
    \scalebox{0.95}{
    \begin{tabular}{cccccc|cc}
    \toprule
       \multirow{2}{*}{Variants} & \multirow{2}{*}{MSSP} & \multirow{2}{*}{SG} & \multirow{2}{*}{GG} & \multicolumn{2}{c|}{VA} & \multirow{2}{*}{IoU (\%)} & \multirow{2}{*}{mIoU (\%)} \\ 
       & & & & GP & OA & & \\
    \midrule
        baseline &  &  &  & &  & 41.76 & 10.62 \\
        1 & $\checkmark$ &  &  &  & & 43.22 & 13.00  \\
        2 & $\checkmark$ & $\checkmark$ &  &  & & 43.32 & 13.68  \\
        3 & $\checkmark$ & $\checkmark$ & $\checkmark$  &  &  & 43.45 & 13.44  \\
        4 & $\checkmark$ & $\checkmark$ & $\checkmark$ & $\checkmark$ &  & 43.14 & 14.39  \\ \hline
        5 & $\checkmark$ & $\checkmark$ & $\checkmark$ & $\checkmark$ & $\checkmark$ & \textbf{43.60} & \textbf{14.55} \\     
    \bottomrule
    \end{tabular}}
    \label{ab:components}
\end{table}

\begin{table}[t]\centering
    \small
    \renewcommand\tabcolsep{5.4pt}
    \renewcommand\arraystretch{1.1}
    \caption{Impact of different training modes. Our one-stage SGN surpasses the two-stage VoxFormer by a large margin. Memory denotes training memory.}
    \scalebox{0.95}{
    \begin{tabular}{lc|cc|cc}
    \toprule
        \multirow{2}{*}{Methods} & \multirow{2}{*}{Mode} & IoU & mIoU & Params & Memory \\
        & & (\%) & (\%) & (M) & (G) \\ 
    \midrule
        VoxFormer-S & two-stage & 44.02 & 12.35 & 57.90 & 14.41\\
        VoxFormer-T & two-state & 44.15 & 13.35 & 57.90 & 16.38\\
        SGN-S (ours) & two-stage & \textbf{44.76} & \textbf{14.93} & \textbf{27.79} & \textbf{10.92} \\\hline
        SGN-S (ours) & one-stage & 43.60 & 14.55 & 28.16 & 14.21\\
        SGN-L (ours) & one-stage & 45.45 & 14.80 & \textbf{12.50} & \textbf{7.16}\\
        SGN-T (ours) & one-stage & \textbf{46.21} & \textbf{15.32} & 28.16 & 15.83\\
    \bottomrule
    \end{tabular}}
    \label{ab:train}
\end{table} 

\begin{table}[t]\centering
    \small
    \renewcommand\tabcolsep{2pt}
    \renewcommand\arraystretch{1.1}
    \caption{Ablation study for depth estimator. Mono and Stereo denote using monocular-based Adabins \cite{bhat2021adabins} and stereo-based MobileStereoNet \cite{shamsafar2022mobilestereonet} as the depth estimator. }
    \scalebox{0.95}{
    \begin{tabular}{c|c|ccc|ccc}
    \toprule
        \multirow{2}{*}{{Methods}} &\multirow{2}{*}{{Depth}} &\multicolumn{3}{c|}{{IoU (\%)}} &\multicolumn{3}{c}{{mIoU (\%)}} \\
        & &{12.8m} &{25.6m} & {51.2m} &{12.8m} &{25.6m} & {51.2m} \\ 
    \midrule
        \multirow{2}{*}{VoxFormer-S} & Mono & 57.41 & 50.61 & 38.68 & 14.62 & 14.01& 10.67\\
        & Stereo & 65.35 & 57.54 & 44.02 & 17.66 & 16.48 & 12.35 \\ \hline
         \multirow{2}{*}{VoxFormer-T} & Mono & 59.03 & 50.47 &  38.08 & 18.67 & 15.42 & 11.27\\
        & Stereo & 65.38 &57.69 &44.15 & 21.55 &18.42 &13.35 \\ \hline
        \multirow{2}{*}{SGN-S (ours)} & Mono & 59.82 & 51.43 & 39.35 & 18.13 & 16.65 & 12.51 \\
         & Stereo &  64.21 & 56.20 & 43.60 & 21.53 & 19.60 & 14.55\\ \hline
        \multirow{2}{*}{SGN-L (ours)} & Mono & 63.79 & 54.99 & 41.36 & 20.77 & 17.69 & 12.64 \\
         & Stereo &  \underline{70.08} & \underline{61.17} & \underline{45.45} & \underline{24.76} & \underline{21.17} & \underline{14.80} \\ \hline
        \multirow{2}{*}{SGN-T (ours)} & Mono & 64.74 & 55.55 & 41.87 & 21.43 & 17.94 & 12.91 \\
        & Stereo &  \textbf{70.61} & \textbf{61.90} & \textbf{46.21} & \textbf{25.70} & \textbf{22.02} & \textbf{15.32}\\
    \bottomrule
    \end{tabular}}
    \label{ab:depth}
\end{table} 

\begin{table}[ht]\centering
    \small
    \renewcommand\tabcolsep{3pt}
    \renewcommand\arraystretch{1.1}
    \caption{Ablation on view transformation. Using FLoSP contains fewer parameters while achieving comparable performance to other variants. Memory denotes training memory.}
    \scalebox{0.95}{
    \begin{tabular}{l|cc|cc}
    \toprule
        Module
        & IoU (\%)
        & mIoU (\%)
        & Params (M)
        & Memory (G) \\
    \midrule
        FLoSP & \underline{43.60} & \underline{14.55} & \textbf{28.16} & \underline{14.21} \\
        LSS & 42.95 & \textbf{14.77} & \underline{30.56} & \textbf{13.23} \\
        Cross-Attention & \textbf{43.66} & 14.05 & 62.15 & 19.01 \\
    \bottomrule
    \end{tabular}}
    \label{ab:transformation}
\end{table} 

\begin{table}[ht]\centering
    \small
    \renewcommand\tabcolsep{4.8pt}
    \renewcommand\arraystretch{1.1}
    \caption{Ablation on image features, including the image backbone and scales. Using scale 16 strikes a balance between performance and model parameters.}
    \scalebox{0.95}{
    \begin{tabular}{c|cccc|ccc}
    \toprule
       \multirow{2}{*}{Backbone} & \multicolumn{4}{c|}{scales} & \multirow{2}{*}{IoU (\%)} & \multirow{2}{*}{mIoU (\%)} & \multirow{2}{*}{Params (M)} \\
        & 4 & 8 & 16 & 32 & & & \\
    \midrule
      \multirow{5}{*}{ResNet50} & $\checkmark$ & & & & 43.10 & 13.85 & \underline{28.06}\\
        & & $\checkmark$ & & & 42.92 & 14.26 & 28.09\\
        & & & $\checkmark$ & & \textbf{43.60} & \underline{14.55} & 28.16 \\
        & & & & $\checkmark$ & 43.36 & 14.20 & 28.29 \\
        & $\checkmark$ & $\checkmark$ & $\checkmark$ & $\checkmark$ & 42.86 & \textbf{14.60} & 28.96 \\
    \hline
       ResNet18 & & & $\checkmark$ & & \underline{43.57} & 14.08 & \textbf{15.73} \\
    \bottomrule
    \end{tabular}}
    \label{ab:image}
\end{table} 


\subsection{Comparison with the state-of-the-art}
\paragraph{SemanticKITTI} Table \ref{tab:semkitti_test} and Table \ref{tab:semkitti_val} present the comparison results between our SGN and other state-of-the-art camera-based SSC methods on the SemanticKITTI validation and test sets, respectively. Our SGN-T achieves state-of-the-art performance on both SemanticKITTI validation and test sets. Specifically, SGN-T outperforms the second one by 1.86\% points (OccFormer) and 2.19\% points (VoxFormer) regarding mIoU and IoU, as shown in Table \ref{tab:semkitti_val}. And compared with these fully dense processing methods with complex 3D models, such as MonoScene and OccFormer, our SGN-S also performs better in terms of mIoU and IoU. For example, SGN-S greatly boosts the MonoScene by 3.05\% points in mIoU and 6.48\% points in IoU, demonstrating the effectiveness of our dense-sparse-dense design equipped with hybrid guidance. Notably, SGN-S outperforms the recent VoxFormer by 2.2\% points in mIoU but has a slightly lower IoU (-0.42\% points). We explain that VoxFormer adopted a two-stage training approach, and the first stage was trained offline with temporal inputs, helping enhance occupancy precision. However, our SGN-S is end-to-end trained with only a single frame as input. 
Compared to the two-stage VoxFormer, the higher mIoU score of our one-stage SGN-S for semantic scene completion demonstrates the superiority of our framework of semantic propagation based on spatial geometry cues.

Remarkably, our lightweight version SGN-L achieves notable performance (45.45\% IoU and 14.80\% mIoU) on SemanticKITTI validation with only \textbf{12.5M} parameters. Compared with MonoScene, OccFormer, and VoxFormer with $\sim$150M, $\sim$200M, and $\sim$60M parameters, our SGN-L performs better while being more lightweight. It demonstrates that our SGN requires no heavy 3D model and has a more powerful representation ability.

\textbf{Quantitative comparison in different ranges.}
We also provide the results of different ranges in Table \ref{tab:comp2lidar}. The results show that SGN-T achieves mIoU scores of 25.70\% and 22.02\% within 12.8 meters and 25.6 meters and performs better than VoxFormer-T by 4.15\% and 3.60\% points in mIoU, respectively. Additionally, our SGN-S surpassed MonoScene by 9.28\% and 7.38\% points in mIoU within 12.8 meters and 25.6 meters. Notably, SGN-T obtains competitive performance with LiDAR-based methods in short-range (12.8 meters) areas. For example, SGN-T outperforms SSCNet by 5.68\% points in mIoU and 6.24\% points in IoU within 12.8 meters,  demonstrating the potential application of our camera-based method for autonomous driving. 

\paragraph{SSCBench-KITTI-360}
Table \ref{tab:kitti_360_test} presents the comparison results between our SGN and other state-of-the-art SSC methods, including LiDAR-based methods (SSCNet, LMSCNet) and camera-based methods (MonoScene, TPVFormer, VoxFormer, OccFormer, DepthSSC, Symphonize) on the SSCBench-KITTI-360 test sets. We can see that our SGN outperforms most camera-based methods by a large margin in terms of both mIou and IoU metrics. For example, SGN-S, SGN-L, and SGN-T surpass OccFormer by 5.95\%, 6.37\%, 6.79\% points in IoU and 3.9\%, 3.14\% 4.44\% points in mIoU. Compared to the LiDAR-based methods, our SGN-T also archives comparable performance in IoU and performs better in mIoU, demonstrating the superiority of our SGN. Interestingly, we found that our SGN-T achieves better performance on many \emph{thing} classes such as traffic-sign, other-object, trucks, bicycles, motorcycles, and other vehicles while performing worse on plain \emph{stuff} classes such as road, parking, sidewalk, building, and vegetation than LiDAR-based method SSCNet. We explain that the LiDAR point cloud contains more accurate structure information, which may facilitate the occupancy prediction of plain classes, while the vision feature includes more semantic information that helps distinguish objects that belong to different classes. 

\begin{figure*}[ht]
\centering
	\includegraphics[width=0.95\textwidth]{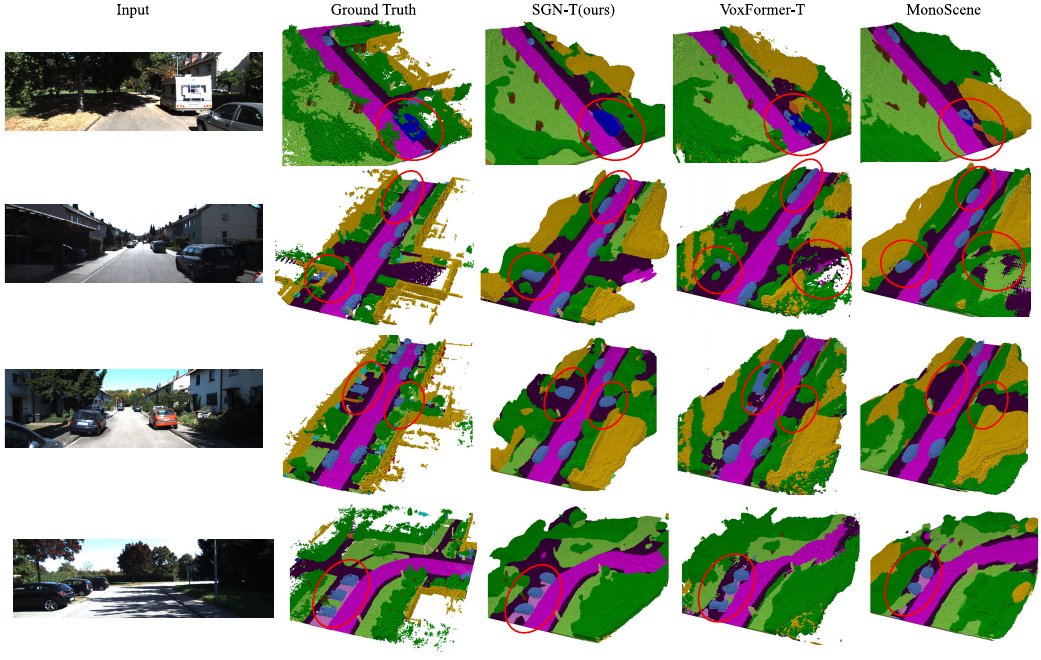}
	\caption{Visual comparison of our SGN-T with state-of-the-art methods on SemanticKITTI validation. Compared to VoxFormer-T and MonoScene, our SGN-T generates more precise segmentation boundaries (labeled in red circles).}
	\label{fig:visualization}
\end{figure*}

\subsection{Qualitative Visualizations}
\begin{figure*}[ht]
\centering
	\includegraphics[width=0.9\textwidth]{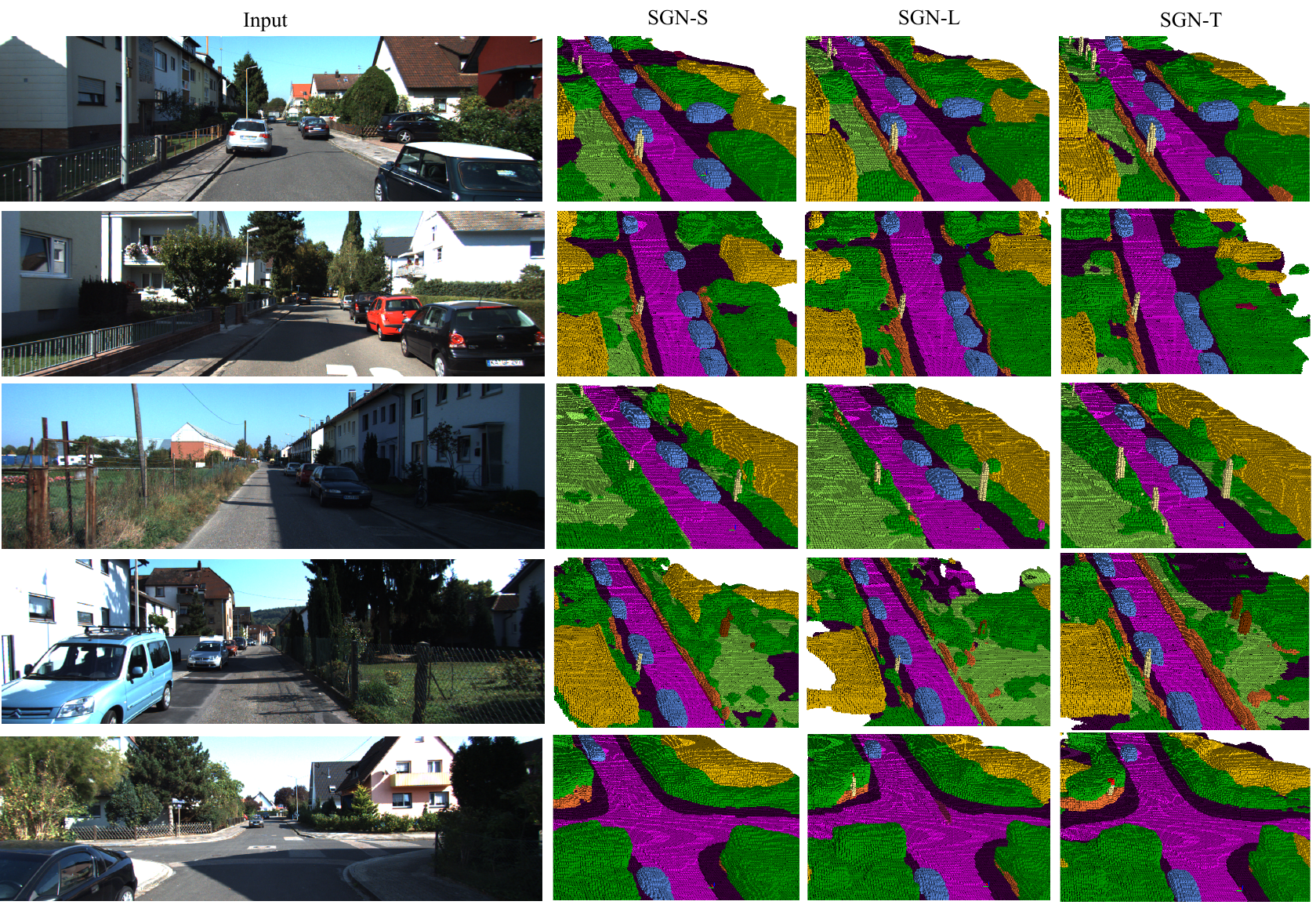}
	\caption{Qualitative results of our SGN-S, SGN-L, and SGN-T on SemanticKITTI hidden test set. Our method can provide accurate semantic occupancy prediction and handle typical driving scenes such as crowded cars, shadows, tiny poles, and crossroads.}
	\label{fig:visual}
\end{figure*}

We provide the visualization results of the proposed SGN-T on SemanticKITTI validation in Figure \ref{fig:visualization}. Compared to VoxFormer-T and MonoScene, our SGN-T generates more precise segmentation boundaries,  especially on ``plane"
classes and large objects such as cars. Besides, SGN-T predicts more accurate SSC results and preserves more regional details in the short-range areas than other methods. For example, there are some wrong semantics and missing objects for VoxFormer-T and MonoScene in the short-range areas, as shown in the first and third rows in Figure \ref{fig:visualization}. However, we noticed that our SGN-T also missed some distant objects that are very small in RGB images. We explain that our SGN uses image features of the 1/16 scale, which may degrade the performance of objects in distant areas. We also provide qualitative results of our SGN-S, SGN-L, and SGN-T on SemanticKITTI hidden test set in Figure 
\ref{fig:visual}. Our method can provide accurate semantic occupancy prediction and road layout and handle typical driving scenes such as crowded cars, shadows, tiny poles, and crossroads.

\subsection{Ablation studies}
We do ablation studies on network components, training mode, depth estimator, image features, view transformation, seed voxels, model dimensions, and temporal input on SemanticKITTI validation. All experiments are conducted with our SGN-S by default.

\paragraph{Ablation on network components} We do ablation studies to analyze the effect of the proposed semantic guidance (SG), geometry guidance (GG), multi-scale semantic propagation (MSSP), and voxel aggregation layer (VA) in Table \ref{ab:components}. GP and OA denote geometry information from 3D features $\textbf{F}^{3D}$ and occupancy-aware features $\textbf{F}_o^{3D}$, respectively. Firstly, we construct a baseline that directly attaches a segmentation head after the selected seed features. As shown in the first line of Table \ref{ab:components}, the constructed baseline has already achieved 41.76\% IoU and 10.62\% mIoU scores, demonstrating our depth-based sparse voxel proposal network can provide effective seed voxels. 
When equipped with our MSSP (Variant 1), the IoU and mIoU scores are improved by 1.46\% points and 2.38\% points, respectively. It demonstrates the effectiveness of multi-scale information propagation.
And our semantic guidance on the seed features further boosts the mIoU by 0.68\% points (Variant 2 vs. Variant 1), showing the importance of intra-category separation of seed features. 
On the other hand, the geometry guidance brings slight improvement in terms of IoU score, while the geometry prior further boosts the mIoU score by 0.95\% points (Variant 4 vs. Variant 3). 
Besides, introducing geometry information in features $F_o^{3D}$ can help improve performance (Variant 5 vs Variant 4). And comparing Variant 5 with Variant 1, the mIoU score is significantly boosted (+1.55\% points), demonstrating the effectiveness of our semantic propagation based on spatial geometry cues.

\begin{table}[t]\centering
    \small
    \renewcommand\tabcolsep{3pt}
    \renewcommand\arraystretch{1.1}
    \caption{Number of model dimensions and depth of MSSP. Memory denotes training memory.}
    \scalebox{0.95}{
    \begin{tabular}{cc|cc|cc}
    \toprule
        Dimensions & Depth & IoU (\%) & mIoU (\%) & Params (M) & Memory (G) \\
    \midrule
        64 & 1 & 43.16 & \textbf{14.24} & \textbf{24.88} & \textbf{7.23}\\
        64 & 2 & \textbf{43.73} & 14.17 & \underline{24.93} & \underline{7.27}\\
        64 & 3 & \underline{43.32} & \underline{14.20} & 24.97 & 8.04\\ \hline
        128 & 3 & \textbf{43.60} & \textbf{14.55} & 28.16 & 14.21\\
    \bottomrule
    \end{tabular}}
    \label{ab:dimension}
\end{table} 

\paragraph{Impact of different training modes} To explore the effect of different training modes, i.e., two-stage and one-stage training, we provide the detailed comparison results with VoxFormer in Table \ref{ab:train}. Line 3 of Table \ref{ab:train} presents the results of our SGN-S with the two-stage training strategy. Note that when equipping SGN-S with a two-stage approach, the first stage remains the same as VoxFormer, and the parameters and training memory of the second stage are calculated for a fair comparison. In the same training configuration, Our SGN-S surpasses VoxFormer-S by a large margin regarding mIoU scores (+2.58\% points). Even our one-stage SGN-S outperforms two-stage VoxFormer-S by 2.2\% points in mIoU. It is worth noting that the model parameters of our SGN-S are only about half of those of VoxFormer-S. For the temporal version, our SGN-T achieves 46.21 IoU and 15.32 mIoU scores, boosting VoxFormer-T by 2.06\% points in IoU and 1.97\% points in mIoU. Our lightweight version, SGN-L, with only 12.5M parameters, also outperforms VoxFormer-T on mIoU and IoU scores while requiring only about half the training memory (7.16 G).

\paragraph{Abaltion on depth estimator} Our sparse voxel proposal network produces the seed voxels based on the depth map predicted by the depth estimator. The generated depth map contains 3D structure information, such as volume surfaces, which has a direct impact on the seed voxel proposal. To quantitatively analyze the impact of the depth estimator, we compare our SGN equipped with the monocular-based AdaBins \cite{bhat2021adabins} and stereo-based MobileStereoNet \cite{shamsafar2022mobilestereonet} with VoxFormer \cite{li2023voxformer}. The results are presented in Table \ref{ab:depth} and show that using the stereo-based depth estimation brought significant performance improvements for both VoxFormer and our SGN, which means a stronger depth estimator that produces more accurate depth maps can boost the performance further. Notably, in the same configurations, our SGN consistently surpasses VoxFormer in all different ranges, demonstrating the effectiveness and superiority of our approaches.

\paragraph{Impact of image features} The 2D features provide a foundation for informative voxel features. We do detailed experiments in Table \ref{ab:image} to explore the impact of the feature scale and image backbone. We see that using 2D features at a scale of 1/16 in ResNet 50 achieves the best IoU score and has comparable mIoU to other variants. And it strikes a balance between performance and model size. The results of using ResNet18 as the backbone are presented in the last line of Table \ref{ab:image} and show that a more lightweight image backbone reduces the model parameters by 12.43M while the performance of the mIoU score drops by 0.47\% points.

\begin{figure}[t]
\centering
	\includegraphics[width=0.49\textwidth]{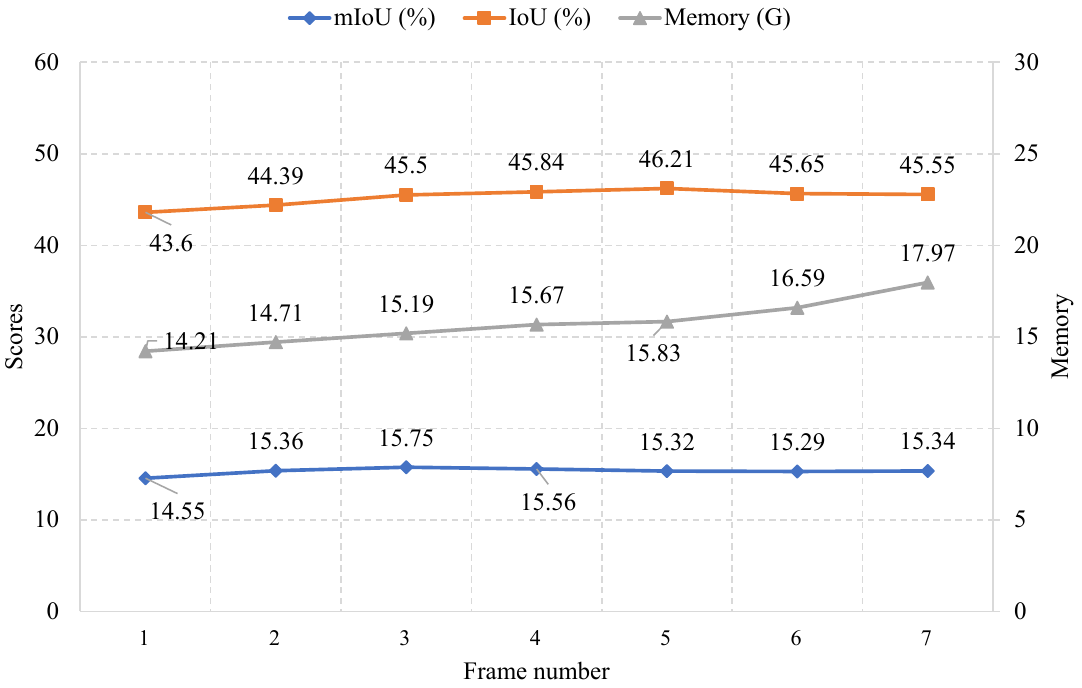}
	\caption{Effect of temporal frames. The frames are sampled every three frames. Memory denotes training memory.}
	\label{fig:mem}
\end{figure}

\paragraph{Ablation on view transformation}
The view transformation generates the initial 3D features for the subsequent hybrid guidance and informative voxel features. We further investigate the effect of different view transformation modules. We implement three commonly used modules, i.e., FLoSP in Monoscene \cite{cao2022monoscene}, LSS \cite{philion2020lift}, and cross-attention adopted from VoxFormer \cite{li2023voxformer}. The results are presented in Table \ref{ab:transformation}, showing that using FLoSP contains fewer parameters while achieving comparable performance to other variants on both mIoU and IoU scores. Therefore, our SGN lifts the 2D features to 3D volume with the view transformation designed in the spirit of FLoSP.

\begin{figure}[t]
\centering
	\includegraphics[width=0.49\textwidth]{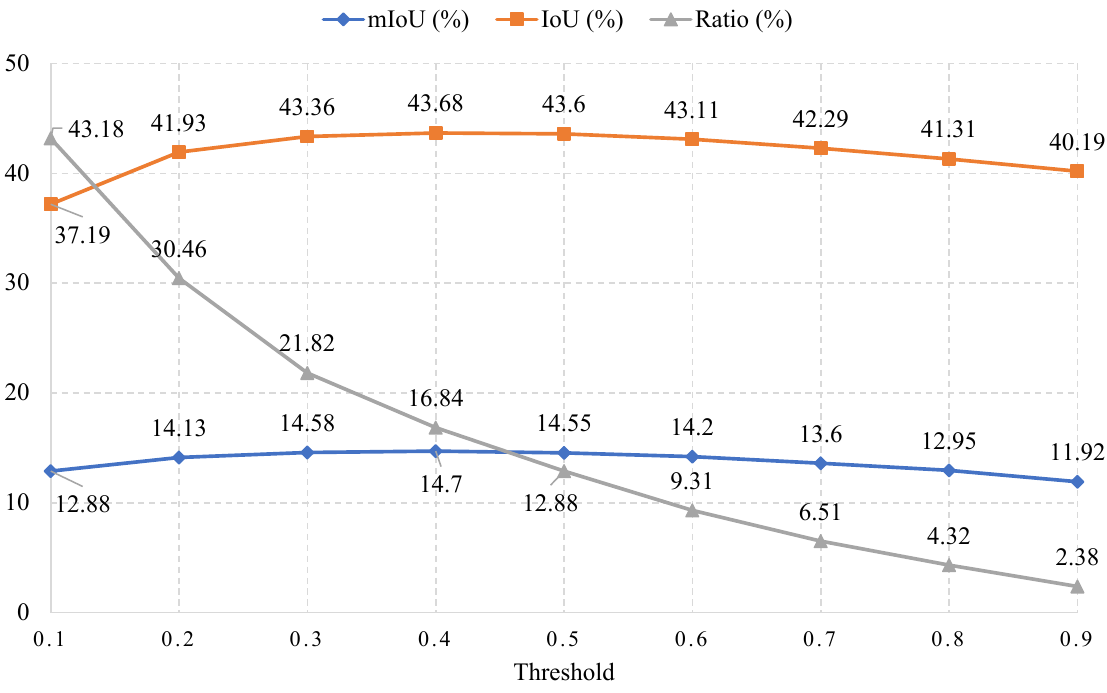}
	\caption{Impact of threshold values for seed voxels. The performance of the model first increases and then decreases as $\theta$ increases. And when $\theta = 0.4$, the model achieves the best performance on both mIoU and IoU scores.  }
	\label{fig:thres}
\end{figure}

\paragraph{Exploration on the threshold for seed voxels}
We change the value $\theta$ to investigate the impact of different thresholds for selecting seed voxels. We calculate the average occupancy rate of seed voxels, mIoU score, and IoU score for variants with $\theta$ from 0.1 to 0.9. The results are shown in Figure \ref{fig:thres}, showing that the performance of the model first increases and then decreases as $\theta$ increases. And when $\theta=0.4$, the model achieves the best performance on both mIoU and IoU scores. Interestingly, we found our model still achieves notable mIoU and IoU scores when the ratio of the seed voxels is very low ($< 5\%$ points). It shows that seed voxels with high confidence play an essential role in our semantic propagation.

\paragraph{Number of model dimensions}
The impact of the number of model dimensions of our SGN-S is evaluated and presented in Table \ref{ab:dimension}. The results reveal that using a large number of feature channels for 3D features boosts the models' performance while increasing the model complexity. For example, the model with 64 dimensions contains fewer parameters and requires less training memory, although its performance drops by 0.28\% points in IoU and 0.35\% points in mIoU (Line3 vs Line4). We also provide the results of different model depths for the multi-scale semantic propagation module. We see that using different depths has comparable mIoU and IoU, demonstrating that our model with hybrid guidance and semantic propagation avoids the dependency on the heavy 3D model for processing 3D features.

\paragraph{Effect of temporal input.}
Finally, we explore the impact of the number of temporal frames in Figure \ref{fig:mem}. We take historical frames to form the temporal input. As Figure \ref{fig:mem} shows, the model's performance on IoU scores first increases with the number of frames and then decreases. We explain that the camera extrinsic matrix from the historical frame to the current system is not always accurate, especially when the temporal interval is long. Therefore, the 3D points may project on the wrong image patches of the historical frames, which disturbs the learning of voxel features. To better balance the performance and memory consumption, our temporal version takes five frames (four past frames and the current frame).

\subsection{Efficiency analysis.}
We perform runtime experiments on a single V100 GPU. The mean value over the SemanticKITTI test set is reported. Our SGN-S, SGN-L, and SGN-T run in 327.71 ms, 315.35 ms, and 436.24 ms, respectively.
We also tested the recent VoxFormer-T (261.46 ms, $\sim$60M parameters) and OccFormer (322.87 ms, $\sim$200M parameters) on the same platform with the officially provided weights for a fair comparison. Compared with these methods, our lightweight version SGN-L achieves better mIoU and IoU scores and comparable latency, but with better applicability due to its lightweight (12.5M parameters) and less training memory consumption (7.16 G).

\newcommand*\rot{\rotatebox{90}}
\newcommand{\ra}[1]{\renewcommand{\arraystretch}{#1}}
\definecolor{LightGrey}{rgb}{.9,.9,.9}
\definecolor{White}{rgb}{1.,0.,1.}
\definecolor{first}{rgb}{.8,.0,.0}
\definecolor{second}{rgb}{.0,.6,.0}
\definecolor{third}{rgb}{.0,.0,.8}
\newcolumntype{g}{>{\columncolor{White}}c}

\definecolor{ceiling}{RGB}{214,  38, 40}   %
\definecolor{floor}{RGB}{43, 160, 4}     %
\definecolor{wall}{RGB}{158, 216, 229}  %
\definecolor{window}{RGB}{114, 158, 206}  %
\definecolor{chair}{RGB}{204, 204, 91}   %
\definecolor{bed}{RGB}{255, 186, 119}  %
\definecolor{sofa}{RGB}{147, 102, 188}  %
\definecolor{table}{RGB}{30, 119, 181}   %
\definecolor{tvs}{RGB}{160, 188, 33}   %
\definecolor{furniture}{RGB}{255, 127, 12}  %
\definecolor{objects}{RGB}{196, 175, 214} %

\definecolor{car}{rgb}{0.39215686, 0.58823529, 0.96078431}
\definecolor{bicycle}{rgb}{0.39215686, 0.90196078, 0.96078431}
\definecolor{motorcycle}{rgb}{0.11764706, 0.23529412, 0.58823529}
\definecolor{truck}{rgb}{0.31372549, 0.11764706, 0.70588235}
\definecolor{other-vehicle}{rgb}{0.39215686, 0.31372549, 0.98039216}
\definecolor{person}{rgb}{1.        , 0.11764706, 0.11764706}
\definecolor{bicyclist}{rgb}{1.        , 0.15686275, 0.78431373}
\definecolor{motorcyclist}{rgb}{0.58823529, 0.11764706, 0.35294118}
\definecolor{road}{rgb}{1.        , 0.        , 1.        }
\definecolor{parking}{rgb}{1.        , 0.58823529, 1.        }
\definecolor{sidewalk}{rgb}{0.29411765, 0.        , 0.29411765}
\definecolor{other-ground}{rgb}{0.68627451, 0.        , 0.29411765}
\definecolor{building}{rgb}{1.        , 0.78431373, 0.        }
\definecolor{fence}{rgb}{1.        , 0.47058824, 0.19607843}
\definecolor{vegetation}{rgb}{0.        , 0.68627451, 0.        }
\definecolor{trunk}{rgb}{0.52941176, 0.23529412, 0.        }
\definecolor{terrain}{rgb}{0.58823529, 0.94117647, 0.31372549}
\definecolor{pole}{rgb}{1.        , 0.94117647, 0.58823529}
\definecolor{traffic-sign}{rgb}{1.        , 0.        , 0.    }   

\makeatletter
\newcommand{\car@semkitfreq}{3.92}
\newcommand{\bicycle@semkitfreq}{0.03}
\newcommand{\motorcycle@semkitfreq}{0.03}
\newcommand{\truck@semkitfreq}{0.16}
\newcommand{\othervehicle@semkitfreq}{0.20}
\newcommand{\person@semkitfreq}{0.07}
\newcommand{\bicyclist@semkitfreq}{0.07}
\newcommand{\motorcyclist@semkitfreq}{0.05}
\newcommand{\road@semkitfreq}{15.30}  %
\newcommand{\parking@semkitfreq}{1.12}
\newcommand{\sidewalk@semkitfreq}{11.13}  %
\newcommand{\otherground@semkitfreq}{0.56}
\newcommand{\building@semkitfreq}{14.1}  %
\newcommand{\fence@semkitfreq}{3.90}
\newcommand{\vegetation@semkitfreq}{39.3}  %
\newcommand{\trunk@semkitfreq}{0.51}
\newcommand{\terrain@semkitfreq}{9.17} %
\newcommand{\pole@semkitfreq}{0.29}
\newcommand{\trafficsign@semkitfreq}{0.08}
\newcommand{\semkitfreq}[1]{{\csname #1@semkitfreq\endcsname}}

\newcommand{\ceiling@nyufreq}{1.37}
\newcommand{\floor@nyufreq}{17.58}
\newcommand{\wall@nyufreq}{15.26}
\newcommand{\window@nyufreq}{1.99}
\newcommand{\chair@nyufreq}{3.01}
\newcommand{\bed@nyufreq}{7.08}
\newcommand{\sofa@nyufreq}{4.70}
\newcommand{\table@nyufreq}{4.31}
\newcommand{\tvs@nyufreq}{0.47}
\newcommand{\furniture@nyufreq}{30.04}
\newcommand{\objects@nyufreq}{14.19}
\newcommand{\nyufreq}[1]{{\csname #1@nyufreq\endcsname}}

\begin{figure*}[t]
\centering
	\includegraphics[width=0.8\textwidth]{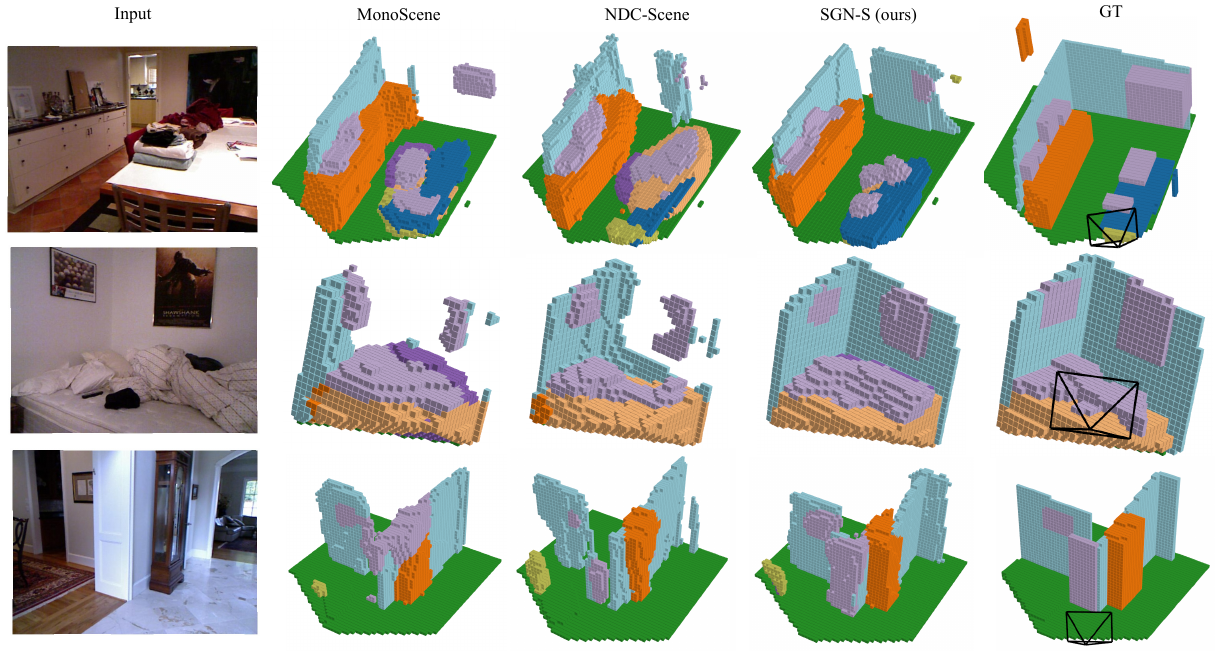}
        \begin{tabular}{c}
             \multicolumn{1}{c}{
				\scriptsize
				\textcolor{ceiling}{$\blacksquare$}ceiling~
				\textcolor{floor}{$\blacksquare$}floor~
				\textcolor{wall}{$\blacksquare$}wall~
				\textcolor{window}{$\blacksquare$}window~
				\textcolor{chair}{$\blacksquare$}chair~
                    \textcolor{bed}{$\blacksquare$}bed~
				\textcolor{sofa}{$\blacksquare$}sofa~
				\textcolor{table}{$\blacksquare$}table~
				\textcolor{tvs}{$\blacksquare$}tvs~
				\textcolor{furniture}{$\blacksquare$}furniture~
				\textcolor{objects}{$\blacksquare$}objects} 
        \end{tabular}
	\caption{Qualitative results of our SGN-S on NYUv2 test set. Our method can provide accurate semantic occupancy prediction, demonstrating the generalization ability on the indoor scenes.}
	\label{fig:nyu}
\end{figure*}

\begin{table*}
    \centering
    \newcommand{\clsname}[2]{
        \rotatebox{90}{
            \hspace{-4pt}
            \textcolor{#2}{$\blacksquare$}
            \hspace{-4pt}
            \renewcommand\arraystretch{0.6}
            \begin{tabular}{l}
                #1~\tiny(\nyufreq{#2}\%) \\
            \end{tabular}
        }}
    \renewcommand{\tabcolsep}{2pt}
    \renewcommand\arraystretch{1.2}
    \caption{\textbf{Semantic Scene completion on NYUv2 test set.} These compared methods are copy from Monoscene \cite{cao2022monoscene} and NDC-scene \cite{yao2023ndc}. Bold and underline denote the best performance and the second-best performance, respectively.}
        \begin{tabular}{l|c|c c c c c c c c c c c|c}
			\toprule
			Method
			& {IoU}
			& \clsname{ceiling}{ceiling}
			& \clsname{floor}{floor}
			& \clsname{wall}{wall} 
			& \clsname{window}{window} 
			& \clsname{chair}{chair} 
			& \clsname{bed}{bed} 
			& \clsname{sofa}{sofa} 
			& \clsname{table}{table} 
			& \clsname{tvs}{tvs} 
			& \clsname{furniture}{furniture} 
			& \clsname{objects}{objects} 
			& mIoU\\
			\midrule

			LMSCNet$^\text{rgb}$~\cite{roldao2020lmscnet} & 33.93 & 4.49 & 88.41 & 4.63 & 0.25 & 3.94 & 32.03 & 15.44 & 6.57 & 0.02 & 14.51 & 4.39 & 15.88 \\
			AICNet$^\text{rgb}$~\cite{li2020anisotropic} & 30.03 & 7.58 & 82.97 & 9.15 & 0.05 & 6.93 & 35.87 & 22.92 & 11.11 & 0.71 & 15.90 & 6.45 & 18.15 \\
			3DSketch$^\text{rgb}$~\cite{chen20203d} & 38.64 & 8.53 & 90.45&	9.94 & 5.67 & 10.64 & 42.29 & 29.21 & 13.88 & 9.38 & 23.83 & 8.19 & 22.91 \\
            
			MonoScene~\cite{cao2022monoscene} & 42.51 & 8.89 & 93.50 &  12.06 &  \underline{12.57} &  13.72 &  48.19 &  36.11 &  \underline{15.13} &  15.22 &  27.96 &  \underline{12.94} &  26.94\\
            NDC-Scene~\cite{yao2023ndc} & 44.17 & 12.02 & 93.51 & 13.11 & \textbf{13.77} & \textbf{15.83} & \textbf{49.57} & \textbf{39.87} & \textbf{17.17} & \textbf{24.57} & \textbf{31.00} & \textbf{14.96} & \textbf{29.03} \\ \midrule
            SGN-S (ours) & \textbf{44.85} & \textbf{14.67} & \textbf{93.56} & \textbf{13.36} & 10.35 & \underline{14.64} & \underline{48.59} &  \underline{37.47} & 14.08 & \underline{15.75} & \underline{30.31} & 12.07 & \underline{27.71} \\
			\bottomrule
		\end{tabular}
		\label{tab:nyu}
\end{table*}

\subsection{Generalization on indoor scenes.}
Although our proposed SGN mainly focuses on the outdoor driving scene as mentioned in Section \ref{method:intro}, we further provide detailed experiments on the NYUv2 \cite{silberman2012indoor} dataset to demonstrate the generalization ability on indoor scenes. NYUv2 comprises 1449 indoor
scenes, represented as $240\times144\times240$ voxel grids labeled with 13 classes (11 semantics, 1 free, 1 unknown). The input resolution is $640\times480$. Following \cite{cao2022monoscene, yao2023ndc}, we utilize a train/test splits of 795/654 scenes to perform evaluations on the test set at the scale of 1:4. Consistent with MonoScene \cite{cao2022monoscene}, to verify the effectiveness on the indoor scenes, we utilize our single-image version SGN-S with the pre-trained EfficientNetB7 \cite{tan1905rethinking} as the image encoder and change the size $X\times Y\times Z$ to $60 \times 36 \times 60$. The 2D feature maps with 1/8 of the input resolution are taken for the subsequent processing. We apply \cite{yang2024depth} to generate monocular depth prediction and train SGN-S for 30 epochs using the AdamW \cite{loshchilov2017decoupled} optimizer with the initial learning rate of 2e-4 and a weight decay of 1e-3. 

As shown in Table \ref{tab:nyu}, without any bells and whistles, our SGN-S achieves the best performance in terms of IoU score and delivers comparable results in mIoU score, which demonstrates the generalization ability of our proposed method on indoor scenes. For instance, SGN-S surpasses MonoScene by 0.77\% and 2.34\% points in mIoU and IoU scores, respectively. However, we observe that NDC-Scene outperforms SGN-S in mIoU score. We attribute this to the complexity and sensitivity of indoor scenes to the accuracy of the depth estimator. Additionally, SGN-S uses an image scale of only 1/8, which may impact the segmentation details, resulting in worse performance on some categories such as ``TVs" and ``objects," as shown in Table \ref{tab:nyu}. We believe that incorporating a more accurate depth estimator, more effective multi-scale feature fusion, and advanced loss designs, such as the frustum proportion loss used in \cite{cao2022monoscene}, could further enhance performance. This will be a focus of our future work. We also present the qualitative results of our method and recent methods on the NYUv2 test set. As illustrated in Figure \ref{fig:nyu}, even without a special design for the indoor scenarios, our SGN-S still generates precise semantic scene completion prediction, which further demonstrates the generalization ability of our method.

\section{Conclusion}
This work focuses on camera-based semantic scene completion (SSC). Existing methods usually rely on sophisticated 3D models to directly process the coarse lifted 3D features that are not discriminative enough for clear segmentation boundaries. Therefore, we propose the one-stage SGN to propagate semantics from the semantic-aware seed voxels to the whole scene based on spatial geometry information. We first redesign the sparse voxel proposal network with the coarse-to-fine paradigm for dynamically and accurately selecting seed voxels. Then, we design hybrid guidance and effective voxel aggregation to enhance the intra-category feature separations and expedite the convergence of semantic propagation. Finally, the multi-scale semantic propagation is proposed for the final semantic scene completion. Extensive experiments on the SemanticKITTI and SSCBench-KITTI-360 benchmarks demonstrate the effectiveness of Our SGN, which achieves state-of-the-art performance while being more lightweight. 

We hope our work can promote the exploration of model optimization and lightweighting in 3D scene understanding and provide innovative solutions for applications in scenarios with limited resources.

\bibliographystyle{IEEEtran}
\bibliography{ref}

\begin{thebibliography}{10}
\providecommand{\url}[1]{#1}
\csname url@samestyle\endcsname
\providecommand{\newblock}{\relax}
\providecommand{\bibinfo}[2]{#2}
\providecommand{\BIBentrySTDinterwordspacing}{\spaceskip=0pt\relax}
\providecommand{\BIBentryALTinterwordstretchfactor}{4}
\providecommand{\BIBentryALTinterwordspacing}{\spaceskip=\fontdimen2\font plus
\BIBentryALTinterwordstretchfactor\fontdimen3\font minus \fontdimen4\font\relax}
\providecommand{\BIBforeignlanguage}[2]{{%
\expandafter\ifx\csname l@#1\endcsname\relax
\typeout{** WARNING: IEEEtran.bst: No hyphenation pattern has been}%
\typeout{** loaded for the language `#1'. Using the pattern for}%
\typeout{** the default language instead.}%
\else
\language=\csname l@#1\endcsname
\fi
#2}}
\providecommand{\BIBdecl}{\relax}
\BIBdecl

\bibitem{roldao2020lmscnet}
L.~Rold{\~a}o, R.~de~Charette, and A.~Verroust-Blondet, ``Lmscnet: Lightweight multiscale 3d semantic completion,'' in \emph{3DV 2020-International Virtual Conference on 3D Vision}, 2020.

\bibitem{cheng2021s3cnet}
R.~Cheng, C.~Agia, Y.~Ren, X.~Li, and L.~Bingbing, ``S3cnet: A sparse semantic scene completion network for lidar point clouds,'' in \emph{Conference on Robot Learning}.\hskip 1em plus 0.5em minus 0.4em\relax PMLR, 2021, pp. 2148--2161.

\bibitem{rist2021semantic}
C.~B. Rist, D.~Emmerichs, M.~Enzweiler, and D.~M. Gavrila, ``Semantic scene completion using local deep implicit functions on lidar data,'' \emph{IEEE transactions on pattern analysis and machine intelligence}, vol.~44, no.~10, pp. 7205--7218, 2021.

\bibitem{yan2021sparse}
X.~Yan, J.~Gao, J.~Li, R.~Zhang, Z.~Li, R.~Huang, and S.~Cui, ``Sparse single sweep lidar point cloud segmentation via learning contextual shape priors from scene completion,'' in \emph{Proceedings of the AAAI Conference on Artificial Intelligence}, vol.~35, no.~4, 2021, pp. 3101--3109.

\bibitem{xia2023scpnet}
Z.~Xia, Y.~Liu, X.~Li, X.~Zhu, Y.~Ma, Y.~Li, Y.~Hou, and Y.~Qiao, ``Scpnet: Semantic scene completion on point cloud,'' in \emph{Proceedings of the IEEE/CVF Conference on Computer Vision and Pattern Recognition}, 2023, pp. 17\,642--17\,651.

\bibitem{li2023lode}
P.~Li, R.~Zhao, Y.~Shi, H.~Zhao, J.~Yuan, G.~Zhou, and Y.-Q. Zhang, ``Lode: Locally conditioned eikonal implicit scene completion from sparse lidar,'' \emph{arXiv preprint arXiv:2302.14052}, 2023.

\bibitem{cao2022monoscene}
A.-Q. Cao and R.~de~Charette, ``Monoscene: Monocular 3d semantic scene completion,'' in \emph{Proceedings of the IEEE/CVF Conference on Computer Vision and Pattern Recognition}, 2022, pp. 3991--4001.

\bibitem{miao2023occdepth}
R.~Miao, W.~Liu, M.~Chen, Z.~Gong, W.~Xu, C.~Hu, and S.~Zhou, ``Occdepth: A depth-aware method for 3d semantic scene completion,'' \emph{arXiv preprint arXiv:2302.13540}, 2023.

\bibitem{wei2023surroundocc}
Y.~Wei, L.~Zhao, W.~Zheng, Z.~Zhu, J.~Zhou, and J.~Lu, ``Surroundocc: Multi-camera 3d occupancy prediction for autonomous driving,'' in \emph{Proceedings of the IEEE/CVF International Conference on Computer Vision}, 2023, pp. 21\,729--21\,740.

\bibitem{zhang2023occformer}
Y.~Zhang, Z.~Zhu, and D.~Du, ``Occformer: Dual-path transformer for vision-based 3d semantic occupancy prediction,'' in \emph{Proceedings of the IEEE/CVF International Conference on Computer Vision}, 2023, pp. 9433--9443.

\bibitem{philion2020lift}
J.~Philion and S.~Fidler, ``Lift, splat, shoot: Encoding images from arbitrary camera rigs by implicitly unprojecting to 3d,'' in \emph{Computer Vision--ECCV 2020: 16th European Conference, Glasgow, UK, August 23--28, 2020, Proceedings, Part XIV 16}.\hskip 1em plus 0.5em minus 0.4em\relax Springer, 2020, pp. 194--210.

\bibitem{li2023voxformer}
Y.~Li, Z.~Yu, C.~Choy, C.~Xiao, J.~M. Alvarez, S.~Fidler, C.~Feng, and A.~Anandkumar, ``Voxformer: Sparse voxel transformer for camera-based 3d semantic scene completion,'' in \emph{Proceedings of the IEEE/CVF Conference on Computer Vision and Pattern Recognition}, 2023, pp. 9087--9098.

\bibitem{li2020anisotropic}
J.~Li, K.~Han, P.~Wang, Y.~Liu, and X.~Yuan, ``Anisotropic convolutional networks for 3d semantic scene completion,'' in \emph{Proceedings of the IEEE/CVF Conference on Computer Vision and Pattern Recognition}, 2020, pp. 3351--3359.

\bibitem{behley2019semantickitti}
J.~Behley, M.~Garbade, A.~Milioto, J.~Quenzel, S.~Behnke, C.~Stachniss, and J.~Gall, ``Semantickitti: A dataset for semantic scene understanding of lidar sequences,'' in \emph{Proceedings of the IEEE International Conference on Computer Vision}, 2019, pp. 9297--9307.

\bibitem{li2023sscbench}
Y.~Li, S.~Li, X.~Liu, M.~Gong, K.~Li, N.~Chen, Z.~Wang, Z.~Li, T.~Jiang, F.~Yu \emph{et~al.}, ``Sscbench: A large-scale 3d semantic scene completion benchmark for autonomous driving,'' \emph{arXiv preprint arXiv:2306.09001}, 2023.

\bibitem{song2017semantic}
S.~Song, F.~Yu, A.~Zeng, A.~X. Chang, M.~Savva, and T.~Funkhouser, ``Semantic scene completion from a single depth image,'' in \emph{Proceedings of the IEEE Conference on Computer Vision and Pattern Recognition}, 2017, pp. 1746--1754.

\bibitem{zou2021up}
H.~Zou, X.~Yang, T.~Huang, C.~Zhang, Y.~Liu, W.~Li, F.~Wen, and H.~Zhang, ``Up-to-down network: Fusing multi-scale context for 3d semantic scene completion,'' in \emph{2021 IEEE/RSJ International Conference on Intelligent Robots and Systems (IROS)}.\hskip 1em plus 0.5em minus 0.4em\relax IEEE, 2021, pp. 16--23.

\bibitem{zhang2018efficient}
J.~Zhang, H.~Zhao, A.~Yao, Y.~Chen, L.~Zhang, and H.~Liao, ``Efficient semantic scene completion network with spatial group convolution,'' in \emph{Proceedings of the European Conference on Computer Vision (ECCV)}, 2018, pp. 733--749.

\bibitem{yang2021semantic}
X.~Yang, H.~Zou, X.~Kong, T.~Huang, Y.~Liu, W.~Li, F.~Wen, and H.~Zhang, ``Semantic segmentation-assisted scene completion for lidar point clouds,'' in \emph{2021 IEEE/RSJ International Conference on Intelligent Robots and Systems (IROS)}.\hskip 1em plus 0.5em minus 0.4em\relax IEEE, 2021, pp. 3555--3562.

\bibitem{mei2023ssc}
J.~Mei, Y.~Yang, M.~Wang, T.~Huang, X.~Yang, and Y.~Liu, ``Ssc-rs: Elevate lidar semantic scene completion with representation separation and bev fusion,'' in \emph{2023 IEEE/RSJ International Conference on Intelligent Robots and Systems (IROS)}.\hskip 1em plus 0.5em minus 0.4em\relax IEEE, 2023, pp. 1--8.

\bibitem{huang2021bevdet}
J.~Huang, G.~Huang, Z.~Zhu, Y.~Ye, and D.~Du, ``Bevdet: High-performance multi-camera 3d object detection in bird-eye-view,'' \emph{arXiv preprint arXiv:2112.11790}, 2021.

\bibitem{wang2022detr3d}
Y.~Wang, V.~C. Guizilini, T.~Zhang, Y.~Wang, H.~Zhao, and J.~Solomon, ``Detr3d: 3d object detection from multi-view images via 3d-to-2d queries,'' in \emph{Conference on Robot Learning}.\hskip 1em plus 0.5em minus 0.4em\relax PMLR, 2022, pp. 180--191.

\bibitem{liu2022petr}
Y.~Liu, T.~Wang, X.~Zhang, and J.~Sun, ``Petr: Position embedding transformation for multi-view 3d object detection,'' in \emph{European Conference on Computer Vision}.\hskip 1em plus 0.5em minus 0.4em\relax Springer, 2022, pp. 531--548.

\bibitem{li2022bevformer}
Z.~Li, W.~Wang, H.~Li, E.~Xie, C.~Sima, T.~Lu, Y.~Qiao, and J.~Dai, ``Bevformer: Learning bird’s-eye-view representation from multi-camera images via spatiotemporal transformers,'' in \emph{European conference on computer vision}.\hskip 1em plus 0.5em minus 0.4em\relax Springer, 2022, pp. 1--18.

\bibitem{wang2021fcos3d}
T.~Wang, X.~Zhu, J.~Pang, and D.~Lin, ``Fcos3d: Fully convolutional one-stage monocular 3d object detection,'' in \emph{Proceedings of the IEEE/CVF International Conference on Computer Vision}, 2021, pp. 913--922.

\bibitem{chen2022polar}
S.~Chen, X.~Wang, T.~Cheng, Q.~Zhang, C.~Huang, and W.~Liu, ``Polar parametrization for vision-based surround-view 3d detection,'' \emph{arXiv preprint arXiv:2206.10965}, 2022.

\bibitem{zhou2022cross}
B.~Zhou and P.~Kr{\"a}henb{\"u}hl, ``Cross-view transformers for real-time map-view semantic segmentation,'' in \emph{Proceedings of the IEEE/CVF conference on computer vision and pattern recognition}, 2022, pp. 13\,760--13\,769.

\bibitem{chen2022efficient}
S.~Chen, T.~Cheng, X.~Wang, W.~Meng, Q.~Zhang, and W.~Liu, ``Efficient and robust 2d-to-bev representation learning via geometry-guided kernel transformer,'' \emph{arXiv preprint arXiv:2206.04584}, 2022.

\bibitem{huang2023tri}
Y.~Huang, W.~Zheng, Y.~Zhang, J.~Zhou, and J.~Lu, ``Tri-perspective view for vision-based 3d semantic occupancy prediction,'' in \emph{Proceedings of the IEEE/CVF Conference on Computer Vision and Pattern Recognition}, 2023, pp. 9223--9232.

\bibitem{jiang2024symphonize}
H.~Jiang, T.~Cheng, N.~Gao, H.~Zhang, T.~Lin, W.~Liu, and X.~Wang, ``Symphonize 3d semantic scene completion with contextual instance queries,'' in \emph{Proceedings of the IEEE/CVF Conference on Computer Vision and Pattern Recognition}, 2024, pp. 20\,258--20\,267.

\bibitem{li2023stereoscene}
B.~Li, Y.~Sun, X.~Jin, W.~Zeng, Z.~Zhu, X.~Wang, Y.~Zhang, J.~Okae, H.~Xiao, and D.~Du, ``Stereoscene: Bev-assisted stereo matching empowers 3d semantic scene completion,'' \emph{arXiv preprint arXiv:2303.13959}, 2023.

\bibitem{yao2023ndc}
J.~Yao, C.~Li, K.~Sun, Y.~Cai, H.~Li, W.~Ouyang, and H.~Li, ``Ndc-scene: Boost monocular 3d semantic scene completion in normalized device coordinates space,'' in \emph{Proceedings of the IEEE/CVF International Conference on Computer Vision}, 2023, pp. 9455--9465.

\bibitem{gan2023simple}
W.~Gan, N.~Mo, H.~Xu, and N.~Yokoya, ``A simple attempt for 3d occupancy estimation in autonomous driving,'' \emph{arXiv preprint arXiv:2303.10076}, 2023.

\bibitem{li2023fb}
Z.~Li, Z.~Yu, D.~Austin, M.~Fang, S.~Lan, J.~Kautz, and J.~M. Alvarez, ``Fb-occ: 3d occupancy prediction based on forward-backward view transformation,'' \emph{arXiv preprint arXiv:2307.01492}, 2023.

\bibitem{tian2023occ3d}
X.~Tian, T.~Jiang, L.~Yun, Y.~Wang, Y.~Wang, and H.~Zhao, ``Occ3d: A large-scale 3d occupancy prediction benchmark for autonomous driving,'' \emph{arXiv preprint arXiv:2304.14365}, 2023.

\bibitem{wang2023openoccupancy}
X.~Wang, Z.~Zhu, W.~Xu, Y.~Zhang, Y.~Wei, X.~Chi, Y.~Ye, D.~Du, J.~Lu, and X.~Wang, ``Openoccupancy: A large scale benchmark for surrounding semantic occupancy perception,'' \emph{arXiv preprint arXiv:2303.03991}, 2023.

\bibitem{bao2019monofenet}
W.~Bao, B.~Xu, and Z.~Chen, ``Monofenet: Monocular 3d object detection with feature enhancement networks,'' \emph{IEEE Transactions on Image Processing}, vol.~29, pp. 2753--2765, 2019.

\bibitem{huang2023obmo}
C.~Huang, T.~He, H.~Ren, W.~Wang, B.~Lin, and D.~Cai, ``Obmo: One bounding box multiple objects for monocular 3d object detection,'' \emph{IEEE Transactions on Image Processing}, vol.~32, pp. 6570--6581, 2023.

\bibitem{cai2023ci3d}
F.~Cai, H.~Chen, and L.~Deng, ``Ci3d: Context interaction for dynamic objects and static map elements in 3d driving scenes,'' \emph{IEEE Transactions on Image Processing}, 2023.

\bibitem{zhang20213d}
W.~Zhang, Y.~Zhang, R.~Song, Y.~Liu, and W.~Zhang, ``3d layout estimation via weakly supervised learning of plane parameters from 2d segmentation,'' \emph{IEEE Transactions on Image Processing}, vol.~31, pp. 868--879, 2021.

\bibitem{li2023bevdepth}
Y.~Li, Z.~Ge, G.~Yu, J.~Yang, Z.~Wang, Y.~Shi, J.~Sun, and Z.~Li, ``Bevdepth: Acquisition of reliable depth for multi-view 3d object detection,'' in \emph{Proceedings of the AAAI Conference on Artificial Intelligence}, vol.~37, no.~2, 2023, pp. 1477--1485.

\bibitem{peng2023bevsegformer}
L.~Peng, Z.~Chen, Z.~Fu, P.~Liang, and E.~Cheng, ``Bevsegformer: Bird's eye view semantic segmentation from arbitrary camera rigs,'' in \emph{Proceedings of the IEEE/CVF Winter Conference on Applications of Computer Vision}, 2023, pp. 5935--5943.

\bibitem{zhang2022beverse}
Y.~Zhang, Z.~Zhu, W.~Zheng, J.~Huang, G.~Huang, J.~Zhou, and J.~Lu, ``Beverse: Unified perception and prediction in birds-eye-view for vision-centric autonomous driving,'' \emph{arXiv preprint arXiv:2205.09743}, 2022.

\bibitem{hu2021fiery}
A.~Hu, Z.~Murez, N.~Mohan, S.~Dudas, J.~Hawke, V.~Badrinarayanan, R.~Cipolla, and A.~Kendall, ``Fiery: Future instance prediction in bird's-eye view from surround monocular cameras,'' in \emph{Proceedings of the IEEE/CVF International Conference on Computer Vision}, 2021, pp. 15\,273--15\,282.

\bibitem{roddick2018orthographic}
T.~Roddick, A.~Kendall, and R.~Cipolla, ``Orthographic feature transform for monocular 3d object detection,'' \emph{arXiv preprint arXiv:1811.08188}, 2018.

\bibitem{liu2022bevfusion}
Z.~Liu, H.~Tang, A.~Amini, X.~Yang, H.~Mao, D.~Rus, and S.~Han, ``Bevfusion: Multi-task multi-sensor fusion with unified bird's-eye view representation,'' \emph{arXiv preprint arXiv:2205.13542}, 2022.

\bibitem{mei2023centerlps}
J.~Mei, Y.~Yang, M.~Wang, Z.~Li, X.~Hou, J.~Ra, L.~Li, and Y.~Liu, ``Centerlps: Segment instances by centers for lidar panoptic segmentation,'' in \emph{Proceedings of the 31st ACM International Conference on Multimedia}, 2023, pp. 1884--1894.

\bibitem{he2016deep}
K.~He, X.~Zhang, S.~Ren, and J.~Sun, ``Deep residual learning for image recognition,'' in \emph{Proceedings of the IEEE conference on computer vision and pattern recognition}, 2016, pp. 770--778.

\bibitem{lin2017feature}
T.-Y. Lin, P.~Doll{\'a}r, R.~Girshick, K.~He, B.~Hariharan, and S.~Belongie, ``Feature pyramid networks for object detection,'' in \emph{Proceedings of the IEEE conference on computer vision and pattern recognition}, 2017, pp. 2117--2125.

\bibitem{geiger2012we}
A.~Geiger, P.~Lenz, and R.~Urtasun, ``Are we ready for autonomous driving? the kitti vision benchmark suite,'' in \emph{2012 IEEE conference on computer vision and pattern recognition}.\hskip 1em plus 0.5em minus 0.4em\relax IEEE, 2012, pp. 3354--3361.

\bibitem{zhou2020cylinder3d}
H.~Zhou, X.~Zhu, X.~Song, Y.~Ma, Z.~Wang, H.~Li, and D.~Lin, ``Cylinder3d: An effective 3d framework for driving-scene lidar semantic segmentation,'' \emph{arXiv preprint arXiv:2008.01550}, 2020.

\bibitem{ye2022efficient}
M.~Ye, R.~Wan, S.~Xu, T.~Cao, and Q.~Chen, ``Efficient point cloud segmentation with geometry-aware sparse networks,'' in \emph{Computer Vision--ECCV 2022: 17th European Conference, Tel Aviv, Israel, October 23--27, 2022, Proceedings, Part XXXIX}.\hskip 1em plus 0.5em minus 0.4em\relax Springer, 2022, pp. 196--212.

\bibitem{shamsafar2022mobilestereonet}
F.~Shamsafar, S.~Woerz, R.~Rahim, and A.~Zell, ``Mobilestereonet: Towards lightweight deep networks for stereo matching,'' in \emph{Proceedings of the IEEE/CVF Winter Conference on Applications of Computer Vision}, 2022, pp. 2417--2426.

\bibitem{ye2021drinet}
M.~Ye, S.~Xu, T.~Cao, and Q.~Chen, ``Drinet: A dual-representation iterative learning network for point cloud segmentation,'' in \emph{Proceedings of the IEEE/CVF international conference on computer vision}, 2021, pp. 7447--7456.

\bibitem{berman2018lovasz}
M.~Berman, A.~Rannen~Triki, and M.~B. Blaschko, ``The lov{\'a}sz-softmax loss: A tractable surrogate for the optimization of the intersection-over-union measure in neural networks,'' in \emph{Proceedings of the IEEE Conference on Computer Vision and Pattern Recognition}, 2018, pp. 4413--4421.

\bibitem{yao2023depthssc}
J.~Yao and J.~Zhang, ``Depthssc: Depth-spatial alignment and dynamic voxel resolution for monocular 3d semantic scene completion,'' \emph{arXiv preprint arXiv:2311.17084}, 2023.

\bibitem{chen2017deeplab}
L.-C. Chen, G.~Papandreou, I.~Kokkinos, K.~Murphy, and A.~L. Yuille, ``Deeplab: Semantic image segmentation with deep convolutional nets, atrous convolution, and fully connected crfs,'' \emph{IEEE transactions on pattern analysis and machine intelligence}, vol.~40, no.~4, pp. 834--848, 2017.

\bibitem{silberman2012indoor}
N.~Silberman, D.~Hoiem, P.~Kohli, and R.~Fergus, ``Indoor segmentation and support inference from rgbd images,'' in \emph{European conference on computer vision}.\hskip 1em plus 0.5em minus 0.4em\relax Springer, 2012, pp. 746--760.

\bibitem{loshchilov2017decoupled}
I.~Loshchilov and F.~Hutter, ``Decoupled weight decay regularization,'' \emph{arXiv preprint arXiv:1711.05101}, 2017.

\bibitem{bhat2021adabins}
S.~F. Bhat, I.~Alhashim, and P.~Wonka, ``Adabins: Depth estimation using adaptive bins,'' in \emph{Proceedings of the IEEE/CVF Conference on Computer Vision and Pattern Recognition}, 2021, pp. 4009--4018.

\bibitem{chen20203d}
X.~Chen, K.-Y. Lin, C.~Qian, G.~Zeng, and H.~Li, ``3d sketch-aware semantic scene completion via semi-supervised structure prior,'' in \emph{Proceedings of the IEEE/CVF Conference on Computer Vision and Pattern Recognition}, 2020, pp. 4193--4202.

\bibitem{tan1905rethinking}
M.~Tan and E.~Le~Q~V, ``rethinking model scaling for convolutional neural networks. 2019,'' 1905.

\bibitem{yang2024depth}
L.~Yang, B.~Kang, Z.~Huang, X.~Xu, J.~Feng, and H.~Zhao, ``Depth anything: Unleashing the power of large-scale unlabeled data,'' \emph{arXiv preprint arXiv:2401.10891}, 2024.

\end{thebibliography}

 \vfill

\end{document}